\documentclass[example,biber]{now-journal} % Replace example with the journal code.
\usepackage{graphicx}
\usepackage{booktabs}
\usepackage{multirow}
\usepackage{tabularx}
\usepackage{colortbl}
\usepackage{xcolor}
\usepackage{makecell}
\usepackage{diagbox}
\usepackage{longtable}
\usepackage{pdflscape}
\usepackage{rotating}   % for sidewaystable
\usepackage{stfloats}
\usepackage{adjustbox}

\usepackage[font=normalsize,labelfont=sf,textfont=sf]{subcaption}

\usepackage{amsmath,amssymb,amsfonts}
\usepackage{mathrsfs}

\usepackage{pifont}
\usepackage{url}
\usepackage{listings}
\usepackage{verbatim}
\usepackage{tcolorbox}
\usepackage{lipsum}
\usepackage{enumitem}

\usepackage{algorithm}
\usepackage{algpseudocode}

\newcommand{\warp}{\boldsymbol{\mathcal{W}}}
\usepackage{amsmath}
\usepackage{amsbsy}
\usepackage{amssymb}
\usepackage{amsfonts}
\usepackage{amsthm}
\usepackage{physics}
\usepackage{datetime}
\usepackage{algorithm}
\usepackage{tcolorbox}
\providecommand{\cref}[1]{Chapter~\ref{#1}}

\providecommand{\fref}[1]{Figure~\ref{#1}}

\providecommand{\R}{\ensuremath{\mathbb{R}}}

\providecommand{\seq}[1]{$\left<#1\right>$}

\renewcommand{\vec}[1]{\ensuremath{\mathbf{#1}}}
\providecommand{\greekvec}[1]{\ensuremath{\boldsymbol{#1}}}
\providecommand{\mat}[1]{\ensuremath{\mathbf{#1}}}

\providecommand{\calA}{\mathcal{A}}

\providecommand{\mZ}{\mat{Z}}

\renewcommand{\va}{\vec{a}}

\providecommand{\ve}{\vec{e}}

\providecommand{\vp}{\vec{p}}

\providecommand{\vx}{\vec{x}}
\providecommand{\vy}{\vec{y}}

\providecommand{\vbeta}{\greekvec{\beta}}

\providecommand{\vrho}{\greekvec{\rho}}

\providecommand{\vtau}{\greekvec{\tau}}
\providecommand{\vphi}{\greekvec{\phi}}

\providecommand{\vpsi}{\greekvec{\psi}}

\tcbuselibrary{theorems}
\newtcbtheorem[]{theo}{Theorem}%
{colback=gray!30,colframe=gray!45!black,fonttitle=\bfseries}{th}
\newtcbtheorem[]{asmp}{Assumption}%
{colback=gray!30,colframe=gray!45!black,fonttitle=\bfseries}{th}
\newtcbtheorem[]{prop}{Property}%
{colback=gray!30,colframe=gray!45!black,fonttitle=\bfseries}{th}
\newtcbtheorem[]{corol}{Corollary}%
{colback=gray!30,colframe=gray!45!black,fonttitle=\bfseries}{th}

\providecommand{\orgname}[1]{#1}
\providecommand{\orgaddress}[1]{#1}
\providecommand{\city}[1]{#1}
\providecommand{\country}[1]{#1}

\definecolor{PastelRed}{HTML}{F8C8C8}
\definecolor{PastelRedFrame}{HTML}{D96A6A}

\definecolor{PastelOrange}{HTML}{F8E0C8}
\definecolor{PastelOrangeFrame}{HTML}{D99A6A}

\definecolor{PastelYellow}{HTML}{F8F0C8}
\definecolor{PastelYellowFrame}{HTML}{D9C96A}

\definecolor{PastelGreen}{HTML}{D8F8C8}
\definecolor{PastelGreenFrame}{HTML}{6AD96A}

\definecolor{PastelTeal}{HTML}{C8F8E8}
\definecolor{PastelTealFrame}{HTML}{6AD9A3}

\definecolor{PastelBlue}{HTML}{C8E0F8}
\definecolor{PastelBlueFrame}{HTML}{6A8DD9}

\definecolor{PastelPurple}{HTML}{E8C8F8}
\definecolor{PastelPurpleFrame}{HTML}{9A6AD9}

\definecolor{PastelPink}{HTML}{F8C8E8}
\definecolor{PastelPinkFrame}{HTML}{D96AB9}

\definecolor{PastelBrown}{HTML}{F0E0D8}
\definecolor{PastelBrownFrame}{HTML}{B7966D}
\newcommand{\revise}[1]{\textcolor{black}{#1}}
\newtcolorbox{taskblock}[2][]{%
  center title,
  fonttitle=\bfseries\footnotesize,
  fontupper=\scriptsize,
  boxrule=0.8pt,
  arc=3pt,
  left=1pt,right=1pt,top=1pt,bottom=1pt,
  title=#2,
  halign=center,
  before skip=4pt,   
  after skip=4pt,    
  colback=blue!5,
  colframe=blue!50!black,
  #1
}

\title{Restore What Matters: Lessons from Joint Restoration and Recognition}

\author[1]{Lanqing Guo$^{\ast}$}
\author[2]{Xijun Wang$^{\ast}$}
\author[3]{Minchul Kim$^{\ast}$}
\author[2]{Yu Yuan}
\author[1]{Wes Robbins}
\author[2]{Xingguang Zhang}
\author[2]{Nicholas Chimitt}
\author[2]{Stanley~H. Chan}
\author[1]{Zhangyang Wang}
\author[3]{Xiaoming Liu}

\affil[*]{These authors contributed equally to this work.}
\affil[1]{\orgname{The University of Texas at Austin},  \orgaddress{\city{Austin}, \country{USA}}}
\affil[2]{\orgname{Purdue University}, \orgaddress{\city{West Lafayette}, \country{USA}}}
\affil[3]{\orgname{Michigan State University},  \orgaddress{\city{Lansing}, \country{USA}}}

\issuevolumeyear{2025}
\issuevolumenumber{xx}
\articledatabox{Received xx xxxxx 2025; Revised xx xxxxx 2025\\[2pt]
ISSN xxxx-xxxx; DOI 10.1561/xxx.xxxxxxxx\\
\copyright\ 2025 xxx}

\keywords{Restoration, Recognition, Physics, Signal Processing, Neuroscience, Biometrics}
\creditline*{Corresponding author: Xiaoming Liu.}

\disclaimer{Your disclaimer goes here.}

\begin{document}

\begin{abstract}
Recognition pipelines typically adopt a restore-then-recognize workflow, yet decades of experience show that generating visually pleasing images seldom translates to improved recognition. We propose a Joint Restoration-for-Recognition (JR\textsuperscript{2}) paradigm: restore only what downstream tasks truly require, with task signals dictating where, how much, and whether restoration is necessary. JR\textsuperscript{2} rests on three pillars: (i) \textit{Physics}, employing optics-accurate turbulence simulation, extensible to blur and noise, to ground restoration in real image formation; (ii) \textit{Neuroscience}, drawing on selective attention and neuroplasticity to direct model capacity toward identity-critical regions and frames while bypassing already-clean inputs; and (iii) \textit{Vision \& Learning}, coupling recognition loss end-to-end through restoration and alignment so that low-level edits maximize high-level identity stability. Evaluations on IARPA-BRIAR show consistent improvements (e.g., TAR@0.01\%FAR +0.6; FNIR@1\%FPIR -2.5), while a quality gate skips $\sim$70\% of clean frames, reducing cost. Ablations confirm physics priors enhance realism, joint training prevents catastrophic forgetting, and selective restoration suffices in many cases. We conclude that better-looking images are neither necessary nor sufficient; restoration modules must be task-driven, selective, and physically aware. Code, pretrained models, and recipes are provided for integration.
\end{abstract}

\section{Introduction}\label{sec:intro}

\subsection{Why Robust Recognition Remains Elusive}
Face, body, and gait recognition systems have attained near-human performance in laboratory settings, fueled by three pillars of progress~\cite{bolme2024data,briar}:
(1) {\em data scale}, exemplified by web-harvested megadatasets such as
MS-Celeb-1M and WebFace260M\;\cite{msceleb,zhu2021webface260m};
(2) {\em architectural innovation}, from
attention-guided backbones\;\cite{shin2022teaching,kprpe} to
margin-based classification heads
(CosFace, ArcFace, CurricularFace, AdaFace)\;\cite{wang2018cosface,deng2019arcface,huang2020curricularface,kim2022adaface};
and
(3) {\em task-specific optimization criteria}, which have steered re-identification (Re-ID) from
short-term person matching\;\cite{wang2018transferable,market}
to cloth-changing, long-term retrieval\;\cite{yang2019person,li2021learning}.
Yet when these models deploy in the wild — peering through kilometer-scale
atmospheric paths, shaky drone footage, or dim security feeds — they lose their
edge.  Recent low-quality imagery benchmarks (IJB-S;\cite{ijbs}, BRIAR;\cite{briar}) report that even state-of-the-art (SOTA) pipelines can suffer {\em double-digit} drops in verification accuracy.

\begin{figure}[h!]
\centering
\includegraphics[width=1.0\linewidth]{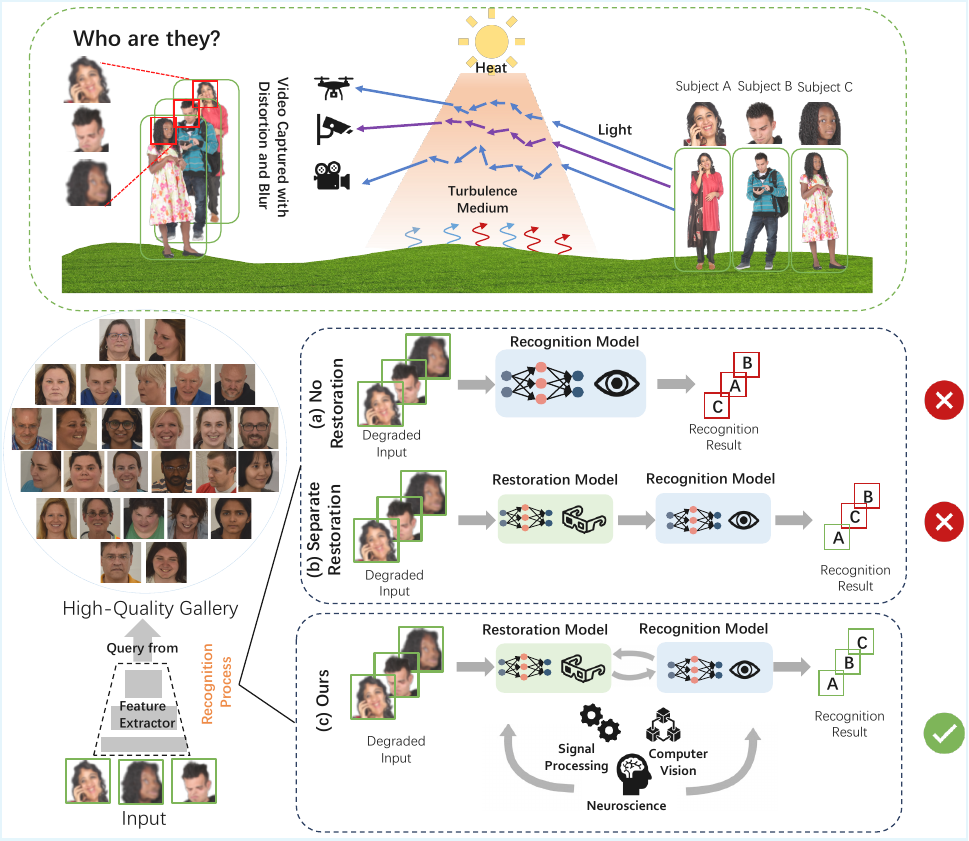}
% \vspace{-5mm}
\caption{\textbf{High-level illustration of the challenges and motivation behind restoration-for-recognition tasks.} (a) The distortion and blur in the degraded input could harm the recognition model a lot. (b)  A two-stage pipeline in which an independent restoration model preprocessed the degraded input before recognition. While this can improve visual quality, it often fails to preserve the identity-specific features needed for accurate recognition. (c) Our joint restoration and recognition framework is driven by interdisciplinary insights from neuroscience, physics-grounded signal processing, and computer vision. \textit{Permission granted by the subject for use of imagery in publications.}}
\label{fig:high-level}
\end{figure}

A traditional remedy is to {\em restore first, recognize later}.
Signal-processing research has delivered powerful tools
for denoising, deblurring, and super-resolution, usually evaluated by
PSNR or SSIM.
Unfortunately, higher perceptual scores do {\em not} guarantee better
downstream recognition:
aggressive denoising can erase micro-textures essential for identity,
whereas mild sharpening often leaves geometric distortions that befuddle
deep matchers.  The community therefore faces a paradox:
\emph{better pixels} do not imply \emph{better semantics}.
As we will also discuss later, our empirical studies on BRIAR dataset, containing long-range atmospheric distortions and significant camera compression, corroborate the dilemma:
naively cascading a turbulence-mitigation network with a state-of-the-art face matcher yields little improvement and even degrades performance on some recognition metrics, whereas a task-aware alternative can achieve gains of over 3\% across all metrics. The high-level illustration of the challenges and motivation behind restoration-for-recognition are shown in Figure \ref{fig:high-level}.

\subsection{An Interdisciplinary Hypothesis}
% Human vision offers a clue.  Neuroscience shows that our cortex uses
% predictive coding, dynamically allocates attention, and adapts mainly in
% higher layers when confronting degraded scenes.  Physics teaches that optical
% wavefront propagation through turbulent media obeys well-characterised
% stochastic operators.  Modern computer vision provides scalable optimisation
% toolkits that can back-propagate task loss through millions of parameters.
% We hypothesise that \emph{melding these three views}
% will yield recognition systems that
% \textbf{restore only what the task needs}.

Human vision provides a biologically grounded blueprint for coping with
imperfect imagery. 
Neuroscience studies of predictive coding posit that higher cortical areas
continually generate hypotheses which are refined by lower–level error
signals, enabling robust perception from noisy or incomplete inputs; this
hierarchical top–down / bottom–up loop has been linked to both endogenous
attention and rapid adaptation in higher visual areas \cite{friston2010free,bastos2012canonical}. Behavioral work further shows that human observers allocate foveal
resources to features carrying the greatest task utility, a mechanism that
parallels the {\em spatial} and {\em temporal} attention blocks now common in
deep backbones \cite{shin2022teaching,kprpe}.  
Such findings imply that, under degradation, a network should {\em not} try
to restore every pixel uniformly but rather devote effort to semantics or identity‐bearing
structures while ignoring background clutter.

Many real–world degradations also lend themselves to
\emph{physics-grounded signal–processing} models.
Atmospheric turbulence, for instance, is well described by Kolmogorov theory:
statistical optics represent the randomly varying phase screen and its
space-variant point-spread function (PSF) through stochastic operators governed by
Fried’s coherence diameter \(r_{0}\) \cite{roddier1981v}.
Embedding such \emph{differentiable} operators inside a learning loop turns
hard domain knowledge into soft, trainable constraints — an idea that has
already boosted physics-guided super-resolution for low-resolution
face recognition \cite{wang2018transferable,8411217,whitelam2017iarpa,klare2015pushing,zha2024dual} and cloth-changing
person Re-ID \cite{yang2019person}.

Finally, modern computer vision supplies the optimization machinery:
end–to–end frameworks can back–propagate biometric losses through millions of
parameters, jointly tuning low‐level filters and high‐level embeddings.
Large–scale datasets such as WebFace260M \cite{zhu2021webface260m} and
MS–Celeb‐1M \cite{msceleb} have demonstrated that, when armed with
task‐appropriate losses (e.g., ArcFace, AdaFace
\cite{deng2019arcface,kim2022adaface}), these networks can absorb vast
quantities of supervision — provided the training data reflect the deployment
domain.

\emph{We therefore hypothesize that converging insights from neuroscience, physics and large‐scale task‐driven learning will yield more robust recognition systems that restore only what the task needs.}

\subsection{A Focused Case Study and Summary of Findings}
\begin{figure}[t]
\centering
\includegraphics[width=1.0\linewidth]{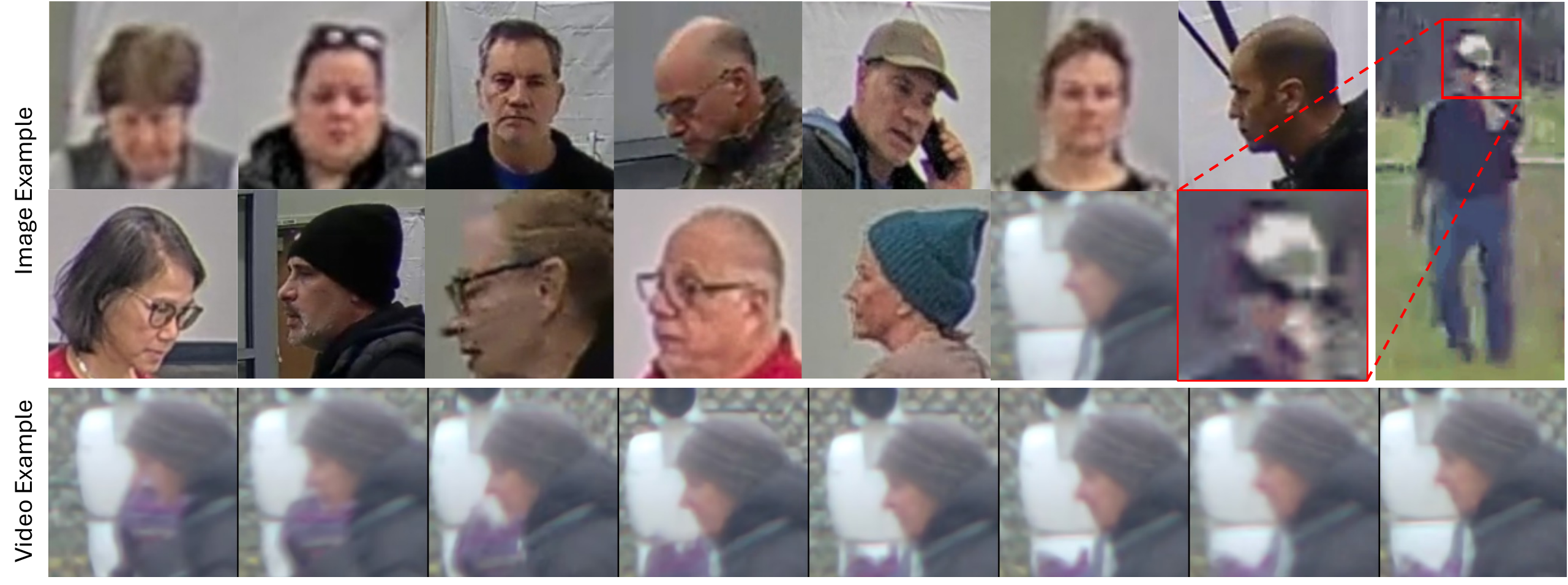}
% \vspace{-5mm}
\caption{\textbf{Examples from the BRIAR dataset.} The top rows show image examples with varying levels of degradation, pose, and occlusion. The bottom row presents a corresponding sequence of video frames. \textit{Permission granted by the subject for use of imagery in publications.}}
\label{fig:briair_example}
\end{figure}

To scrutinize the hypothesis, we target one practically urgent
but scientifically rich degradation: \textbf{atmospheric turbulence}.
Turbulence is a compound nuisance: it simultaneously blurs, warps, and
spatially modulates noise, making it an ideal stress-test for restoration
techniques. 
The public BRIAR corpus (see examples in Figure~\ref{fig:briair_example}), developed under the IARPA program to support biometric recognition in challenging conditions, offers kilometer-scale video with identity labels from aerial and ground views, enabling rigorous evaluation under real-world distortions like atmospheric turbulence.

% The public BRIAR corpus provides kilometer-scale data with identity labels, enabling large-scale, rigorous evaluation. 

We distill the paper’s findings into five concrete contributions, each rooted in - and made possible by - an explicit fusion of \textit{physics}, \textit{neuroscience}, and \textit{computer vision}:

\begin{enumerate}
\item \textbf{Interdisciplinary \emph{Tri-Pillar} pipeline.}  
      We are the first to couple (i) an optics-accurate turbulence simulator (PATS-v5) \cite{roddier1981v,chimitt2024scattering} into (ii) a neuroscience-inspired, attention-guided restoration network \cite{friston2010free,bastos2012canonical} and (iii) a task-aware co-optimization loop with differentiable alignment \cite{wang2018cosface,deng2019arcface}, back-propagating biometric information through \emph{each} component of the pipeline.

\item \textbf{Systematic dissection of the restore–recognize trade-off.}  
      Using the kilometer-scale IARPA–BRIAR benchmark, we conduct the largest ablation study to date:  
      \emph{physics priors} (PATS-v5 vs. PATS-v3) add +1.4\,pp Rank-20 accuracy,  
      and \emph{joint optimization} lowers FNIR@1\%FPIR from 52.4\% to 49.9\%.  

\item \textbf{Quality-gated selective processing.}  
      A lightweight Swin-Transformer classifier bypasses restoration on 70\% of already-clean videos, saving inference FLOPs while \emph{improving} TAR@0.1\%FAR from 62.6\% to 63.3\% and reducing FNIR@1\%FPIR to 49.9\%. We pinpoint when restoration \emph{helps}, \emph{hurts}, or is \emph{unnecessary}.

\item \textbf{Actionable guidelines for task-aware restoration.}  
      Our study consolidates best practices: \emph{embed physics, attend to identity cues, co-optimize with the task, fine-tune late layers, and skip restoration when safe}.  
      These prescriptions translate interdisciplinary insight into a practical design blueprint for future robust recognition systems.

    \item \textbf{ Reproducible resources for the community}. 
      We will release (i) the PATS-v5 turbulence simulator, (ii) 1.8k synthetic turbulence videos paired with clean ground truth, (iii) trained weights, and (iv) full evaluation scripts - lowering the barrier for cross-disciplinary research at the signal-processing/vision interface.
\end{enumerate}

\section{Literature Review}\label{sec:relatedwork}

% \subsection{Restoration as Preprocessing}

% \subsection{Multi-Task Learning}

% \subsection{}

% The intersection of image restoration and recognition has attracted growing interest. However, many existing approaches appear to underutilize interdisciplinary insights from fields such as neuroscience, signal processing, and computer vision. 
In this section, we review related work across interdisciplinary domains—including computer vision, neuroscience, and signal processing—to discuss prior explorations in image restoration and recognition, highlight existing gaps, and motivate the need for a unified, task-aware framework.

% In this section, we review related work across interdisciplinary domains from computer vision, neuroscience foundations, signal processing to review existing 
% exploration on image restoration and recognition and
% identify existing gaps and motivate the development of a more unified, task-aware framework.

\subsection{Computer Vision for Restoration + Recognition}

% Face, body, and gait recognition have significantly advanced over the past decade, driven by the availability of larger datasets, more sophisticated model architectures, and specialized loss functions. In face recognition (FR), massive datasets such as WebFace260M and MS-Celeb-1M~\cite{zhu2021webface260m, msceleb}, along with attention-based models~\cite{shin2022teaching,kprpe} and margin-based softmax losses~\cite{wang2018cosface, deng2019arcface, huang2020curricularface, kim2022adaface}, have contributed to remarkable accuracy in controlled settings. Similarly, body recognition, commonly referred to as person re-identification (ReID), has progressed from short-term matching~\cite{wang2018transferable, li2018unsupervised, li2019unsupervised, lin2019bottom,yu2019unsupervised, zhai2020ad,ge2020mutual,ge2020self,market} to long-term, cloth-changing ReID~\cite{yang2019person, li2021learning, chen2021learning,hong2021fine,gu2022clothes,jin2022cloth,wan2020person,yu2020cocas}, driven by advancements in computer vision and deep learning. Collectively, these advancements have propelled recognition accuracy under relatively ideal scenarios, yet real-world degradations such as low resolution and atmospheric turbulence continue to pose formidable challenges as evidenced by low recognition performance in benchmarks such as IJB-S~\cite{ijbs} and BRIAR dataset~\cite{briar}. Therefore, utilizing restoration techniques with recognition task is indispensable for achieving reliable real-world performance.

Computer vision approaches to image restoration~\cite{rudin1992nonlinear,figueiredo2007gradient,choi2020statnet,jiu2021deep,lau2021semi,LpezTapia2023VariationalDA, gao2023implicit,bouchard2023resolution,you2024indigo+} have historically prioritized improving perceptual fidelity, typically reported with metrics such as Peak Signal-to-Noise Ratio (PSNR)~\cite{psnr} and the Structural Similarity Index Measure  (SSIM)~\cite{wang2004image}. These scores measure visual similarity to a reference (ground-truth) image. Yet numerous studies have shown that higher PSNR/SSIM does not necessarily yield better downstream performance in perception, classification, or recognition tasks~\cite{zhu2012context,blau2018perception,tzirakis2017end,wang2021towards}. When the end goal is recognition, the gold-standard evaluation is the recognition task itself---for example, measuring mean Average Precision (mAP) after enhancement in detection/recognition pipelines~\cite{chen2021pre}, or reporting face identification/verification accuracy after applying turbulence-mitigation restoration~\cite{zhang_2024_TMT}. This disconnect between perceptual quality and task utility motivates research that explicitly bridges restoration and recognition~\cite{chen2021pre,zhang_2024_TMT}.
Importantly, enhanced perceptual quality does not necessarily correlate with improved performance on downstream tasks. Recognition research increasingly explores ways to bridge restoration and recognition: 

% Computer Vision approaches to image restoration~\cite{rudin1992nonlinear,figueiredo2007gradient,choi2020statnet,jiu2021deep,lau2021semi,gao2023implicit,you2024indigo+} have historically prioritized enhancing perceptual fidelity, often evaluated using metrics such as PSNR~\cite{psnr} and SSIM~\cite{wang2004image}. These metrics quantify visual closeness to ground-truth. However, improvements measured by them do not reliably translate into better performance for downstream tasks such as visual perception, classification or recognition \cite{zhu2012context,blau2018perception,tzirakis2017end,wang2021towards}.  Ultimately, the most compelling evaluation involves measuring direct impact on the downstream application itself, i.e., assessing mAP after image enhancement~\cite{chen2021pre}, or quantifying face recognition accuracy when applying restoration techniques like turbulence mitigation~\cite{zhang_2024_TMT}. Importantly, enhanced perceptual quality does not necessarily correlate with improved performance on downstream tasks. Recognition research increasingly explores ways to bridge restoration and recognition: 

\begin{figure}[t]
\centering
\includegraphics[width=1.0\linewidth]{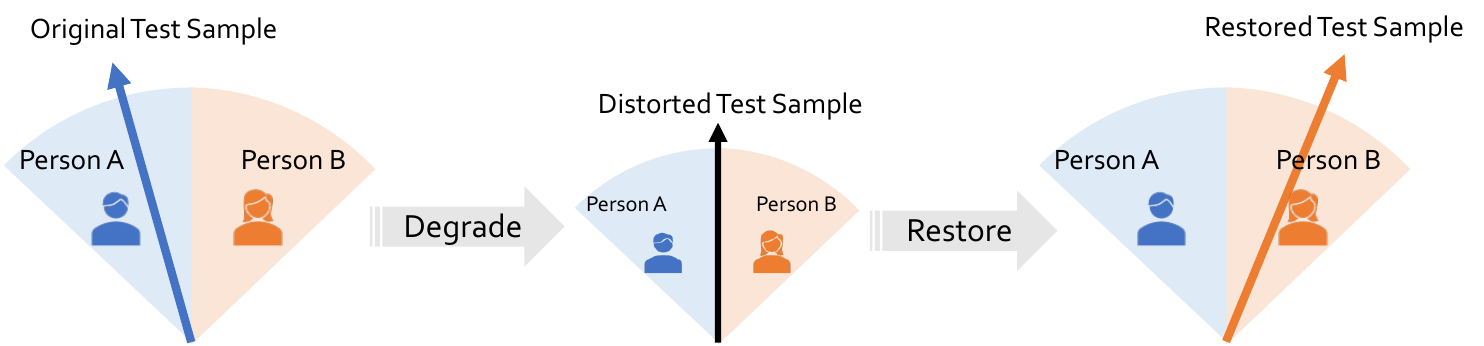}
\vspace{-5mm}
\caption{\textbf{Why restoration may hurt recognition?} Whenever restoration is performed, the prior used in the restoration has already implicitly introduced bias. This bias, if added wrongly, can lead to false recognition in downstream.}
\label{fig:restore_only_paradox}
\end{figure}

\begin{itemize}

    \item Many existing methods~\cite{Jin2021NeutralizingTI, nah2017deepdeblur, anantrasirichai2013atmospheric, jaiswal2023physics, mao2022single, Wang2023AtmosphericTC, mei2023ltt} treat restoration as a pre-processing step, independently cascading trained restoration and recognition models. However, this strategy is often suboptimal, as restoration does not consistently improve recognition accuracy. A key reason is that the restoration process implicitly relies on data-driven priors such as smoothness, facial symmetry, or typical facial feature locations learned from training data~\cite{nah2017deepdeblur}. These priors, shaped by the training distribution, may be inaccurate or misaligned with the actual identity in specific cases. As a result, the restored image can distort identity cues and degrade downstream recognition performance (\fref{fig:restore_only_paradox}).

    \item More sophisticated approaches investigate joint training schemes\cite{tu2021joint, yang2023visual}, end-to-end optimization \cite{chen2021pre}, or task-aware restoration objectives \cite{zhang_2024_TMT, yang2024irvr}. These methods explicitly aim to optimize the restoration process not just for visual quality, but to specifically enhance features relevant to the subsequent recognition task. 
\end{itemize}

Nevertheless, achieving this synergy effectively remains a challenge in the literature. Since there is no guarantee that restoration will preserve the identity, it is necessary to co-optimize the recognition and restoration. To this end, this work presents the first attempt to integrate dynamic attention mechanisms and neuroplasticity-inspired insights from neuroscience with physics-based degradation modeling from signal processing, aiming to establish a robust interdisciplinary framework for the restore-recognize problem.

\vspace{-1mm}
\subsection{Neuroscience Foundations}
\noindent\textbf{Visual Perception Under Adverse Conditions.} The human brain’s ability to recognize objects under severe degradations provides valuable insights for designing robust Artificial Intelligence (AI) systems. Mechanisms such as amodal completion, which enables the inference of missing parts of objects, and predictive coding, where higher-level areas anticipate sensory inputs, are particularly relevant. These capabilities allow humans to recognize occluded or noisy images by integrating hierarchical cues and leveraging contextual information. Research in neuroscience has demonstrated the role of hierarchical processing and attention mechanisms in focusing on critical features under uncertainty \cite{himberger2018principles, itti2002model, gold2017visual}.

Most restoration pipelines lack dynamic, hierarchical mechanisms to prioritize task-relevant features. Integrating neuroscience-inspired designs, such as predictive attention and feedback loops, could bridge this gap. Our experimental findings demonstrate the importance of similar hierarchical reconstruction mechanisms in machine models. For instance, \textit{attention-driven co-optimization of restoration and recognition pipelines} led to significant gains in degraded scenarios.

\vspace{1mm}
\noindent\textbf{Dynamic Attention and Contextual Prioritization. }The brain uses a dynamic interplay of bottom-up and top-down attention mechanisms to selectively focus on task-relevant features. Bottom-up processes respond to salient inputs, while top-down processes guide attention based on task goals and prior knowledge \cite{gold2017visual}. In neuroscience-inspired systems, region-wise prioritization during restoration can mirror these mechanisms. Experimental evidence from our turbulence datasets reveals that guiding restoration efforts toward facial features and learning to focus from recognition yields higher recognition accuracy.
% yields up to 20\% higher recognition accuracy, especially in face biometrics.

\vspace{1mm}
\noindent\textbf{Neuroplasticity.} The brain’s capacity to adapt to recurring or evolving conditions offers a compelling parallel for tuning AI systems \cite{diniz2023times}. In the context of degraded inputs, neuroplasticity enables the brain to recalibrate its processing pipelines, selectively optimizing certain pathways to enhance performance in specific tasks. Translating this concept, we explored \textit{targeted fine-tuning }of AI models to determine which components of the restoration-recognition pipeline are most impactful under a wide range of degradation conditions. Our results show that \textit{selectively tuning} specific layers in the restoration module, guided by recognition feedback, yields substantial gains, suggesting a \textit{plasticity-inspired approach} to pipeline tuning, where models are designed to dynamically adjust specific layers based on degradation-specific challenges.
% For instance:
% \begin{itemize}
%     \item \textbf{Early Layers}: Fine-tuning early layers of the pipeline proved effective for handling low-level degradations like noise and blur. These layers were critical for feature extraction and noise suppression, leading to a \textbf{10\% increase in recognition accuracy} under Gaussian noise.
%     \item \textbf{Middle Layers}: Adjusting mid-layer parameters, which interact with task-specific degradation priors (e.g., turbulence modeling), had the largest impact on scenarios involving complex atmospheric conditions. This yielded \textbf{up to 12\% accuracy improvements} in face and gait recognition tasks.
%     \item \textbf{Late Layers}: Tuning late layers, which directly influence high-level semantic feature alignment, was particularly effective for restoring identity-preserving features in face recognition, boosting accuracy by \textbf{15\% in multi-modal datasets}.
% \end{itemize}
% These results suggest a \textbf{plasticity-inspired approach} to pipeline tuning, where models are designed to dynamically adjust specific layers based on the type and severity of degradation.

\vspace{1mm}
\noindent\textbf{Predictive Coding in Temporal Restoration.} For video data, the brain anticipates future frames based on motion dynamics, maintaining coherence across time \cite{millidge2024predictive}. Inspired by this, predictive coding principles were integrated into \textit{spatiotemporal restoration models} in our study \cite{liu2024farsight}, reducing recognition errors caused by temporal inconsistencies in gait recognition tasks. This highlights the importance of integrating temporal restoration with adaptive tuning mechanisms.

% by over \textbf{15\% in gait recognition tasks}

\subsection{Signal Processing in Turbulence}
% Mention classical restoration approaches (Wiener filters, wavelet-based denoisers, etc.) and highlight their strengths/limitations for task-agnostic settings.
Recent works have explored combining signal priors with deep learning~\cite{Wang2020DeepLearningTomographic,huang2023self,dehner2023deep,oppliger2024weak,zhang2024reusability}, such as embedding wavelet-based sparsity \cite{hsu2023wavelet} or total variation \cite{huo2024image} constraints into neural networks. Despite their promise, these hybrid methods rarely account for task-specific features essential for recognition. Bridging this gap requires integrating physics-driven priors with task-aware co-optimization.

\vspace{1mm}
\noindent\textbf{Physics-Based Signal Simulation and Modeling.} Signal processing provides foundational tools for modeling complex degradations, particularly those driven by physical phenomena such as turbulence, motion blur, and noise. 
Physics-based models like split-step methods \cite{zhu2012removing} which numerically propagate waves through phase screens that represent the atmosphere's spatially varying index of refraction. These simulators enable controlled experimentation and algorithm development by providing realistic approximations of real-world conditions.
Our efforts extend these principles by embedding physics-driven simulators directly into restoration loops. For example:
The Purdue Atmospheric Turbulence Simulator (PATS) \cite{chimitt2020simulating, chan_turbulence, Mao_2021_a, chimitt2022real} was integrated with our restoration networks to generate synthetic turbulence degradations tailored to specific environmental conditions.
% By combining PATS with domain-specific degradations, such as high-altitude or urban turbulence profiles, we achieved 3-5\% recognition accuracy improvements in facial biometrics and text recognition over baseline restoration approaches.
This integration allows restoration models to ``understand'' the degradation processes they are tasked to mitigate, enabling more precise feature preservation during the restoration phase.

\vspace{1mm}
\noindent\textbf{Physics-Based Simulation in the Loop for Robust Training.} Training robust AI systems requires diverse and realistic training data, particularly for complex degradations like turbulence or non-uniform noise. Physics-based simulators play a crucial role in generating such datasets\cite{jaiswal2023physics, Mao_2021_a, mao2022single, zhang_2024_TMT}. In our pipeline:
synthetic degradation data generated using PATS and Kolmogorov models enriched training datasets for turbulence restoration, resulting in lower generalization errors.

\section{Background and Analysis}\label{sec:preliminary}
Image degradation, like atmospheric turbulence, due to adverse environmental effects can generally be described via a generic forward model
\begin{equation}
\vy = \calA(\vx) + \ve,
\end{equation}
where \(\mathcal{A}(\cdot)\) is a non-linear operator describing how the high-quality video clip \(\mathbf{x} \in \mathbb{R}^{H \times W \times F}\) is distorted to produce an observed counterpart \(\mathbf{y} \in \mathbb{R}^{H \times W \times F}\), with \(\mathbf{e} \sim \mathcal{N}(0, \sigma^2 \mathbf{I})\) modeling additive noise. Here, \(H\) and \(W\) denote the spatial height and width of each frame, and \(F\) is the total number of frames in the clip.
% where $\calA(\cdot)$ is a non-linear operator describing how the high-quality video clip $\vx \in \R^{H\times W\times F}$ is distorted to produce an observed counterpart $\vy \in \R^{H\times W\times F}$, with $\ve \sim \calN(0,\sigma^2\mI)$ being a noise term to model any error. 
Since recognition models are generally designed and trained on clean images, the common wisdom is to consider a sequential operation:
\begin{equation}
\arg\min_{\theta, \varphi} \mathcal{G}_{\varphi}(\mathcal{R}_{\theta}(\mathbf{y})),
\end{equation}
by training the restoration network $\mathcal{R}_{\theta}$ and recognition network $\mathcal{G}_{\varphi}$. 
% Because of the embedded nature of the restoration \emph{inside} the recognition, people often separate the two problems and train networks individually. 
We consistently employ the restoration module GRTM\cite{liu2024farsight} together with the recognition module AdaFace\cite{kim2022adaface} across all following experiments. For evaluation, we construct a diverse test set from BRIAR and assess recognition performance using standard metrics: Rank-20, TAR@FAR=0.001 (True Accept Rate at False Accept Rate 0.1\%, which reflects verification accuracy under strict false match control), TPIR@FNIR=0.01 (True Positive Identification Rate at False Negative Identification Rate 1\%), and their mean. Further details on the evaluation metrics are provided in Section~\ref{sec:evaluation}.

Due to the embedded nature of restoration \emph{within} the recognition pipeline, the two tasks are often treated separately and optimized independently.
Such decoupled training typically leads to sub-optimal recognition performance, as the restoration module focuses solely on image quality, while the recognition model is trained for face identification.

\begin{figure}[t]
\centering
\includegraphics[width=.8\linewidth]{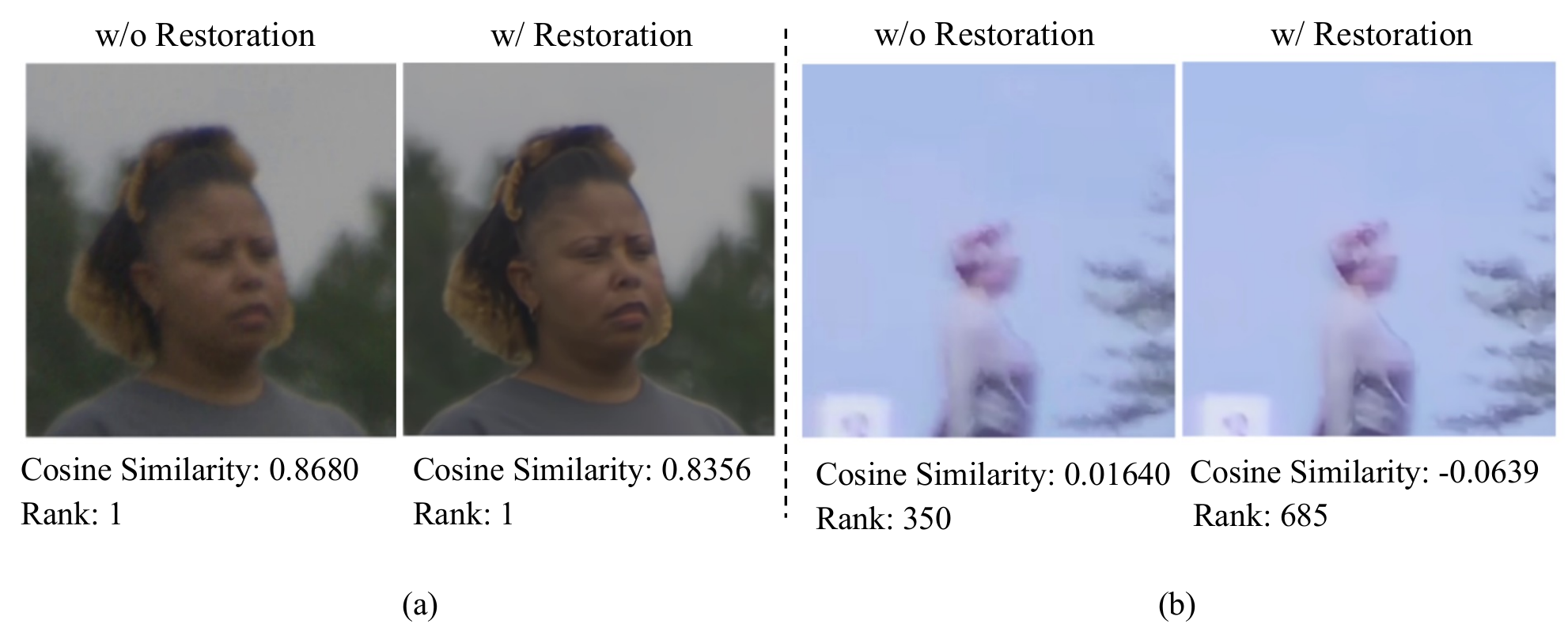}
% \includegraphics[height=4.5cm]{./pix/img_rk_b}
% \vspace{-2mm}
\caption{\textbf{Restoration does not always improve recognition.} (a): while the image quality is clearly improved, the similarity score actually drops. Please zoom in for best visual experience. (b): even though there is no noticeable change in terms of image quality, the recognition rate drops significantly after restoration. Please zoom in for best visualization. \textit{Images shown
with subject permission for publication.}}
\label{fig:example}
\end{figure}

\fref{fig:example} shows two representative cases comparing recognition results with and without restoration as preprocessing, highlighting the following observations:
\begin{itemize}
\item \fref{fig:example}(a): While image visual quality improves, the similarity score decreases — indicating that improved restoration does not necessarily lead to better recognition.
\item \fref{fig:example}(b): Image quality remains unchanged, and the recognition rank drops significantly. Artifacts introduced by restoration adversely affect recognition performance.
\end{itemize}

% To further quantify this effect, \fref{fig:metric_relation} examines the correlation between image quality metrics (PSNR, SSIM, LPIPS, FID) and identity similarity (cosine score). Results on a synthetic BRIAR gallery dataset with 1782 images and 180 identities degraded by simulated turbulence show that high perceptual scores do not imply better recognition. In some cases, images with the high PSNR or SSIM still yield poor identity similarity.

% \fref{fig:metric_relation} further quantitatively analyzes the relationship between image quality metrics and identity recognition, we visualize the correlation between commonly used image quality metrics — PSNR, SSIM, LPIPS, and FID — and the identity preservation metric, cosine similarity. The data is derived from restored images produced by different image restoration models on a synthetic test set from BRIAR gallery degraded by simulated atmospheric turbulence, which contains 1782 face images with 180 unique person IDs. Our observations further indicate that higher image quality metrics do not correspond to better recognition performance. For instance, a degraded input may exhibit the highest PSNR or SSIM values while yielding a low cosine similarity score.

% \begin{figure}[h]
% \centering
% \includegraphics[width=1.\linewidth]{./pix/psnr_ssim_compare.png}
% \caption{The relationship between image quality metrics, measured by PSNR, SSIM, LPIPS, and FID and recognition performance measured by cosine similarity (COS). PSNR, SSIM, and cosine similarity are higher the better, LPIPS and FID are lower the better. Each subplot visualizes the correlation between one image quality metric and cosine similarity across different image restoration models. Each point represents a restoration model, PiRN\cite{jaiswal2023physics}, CodeFormer\cite{zhou2022towards}, GFPGAN\cite{wang2021towards}, AT-DDPM\cite{nair2023ddpm}, DR2\cite{wang2023dr2}, and RTM+\cite{zhang2024spatio} are used. }
% \label{fig:metric_relation}
% \end{figure}

% \fref{fig:metric_relation} further visualizes the relationship between image quality metrics and the identity preservation metric.
% Again, these quantitative analyzes indicate that a better image quality score does not necessarily lead to better recognition accuracy.

\begin{tcolorbox}[before skip=2mm, after skip=0.0cm, boxsep=0.0cm, middle=0.0cm, top=0.1cm, bottom=0.1cm]
    \textit{\textbf{Findings:}\\
    \textit{Better image $\neq$ better recognition}; in some cases, restoration can even degrade recognition performance.
    }
\end{tcolorbox}
\vspace{3mm}

\begin{figure*}[h]
\centering
% \vspace{-5mm}
\includegraphics[width=1.\linewidth]{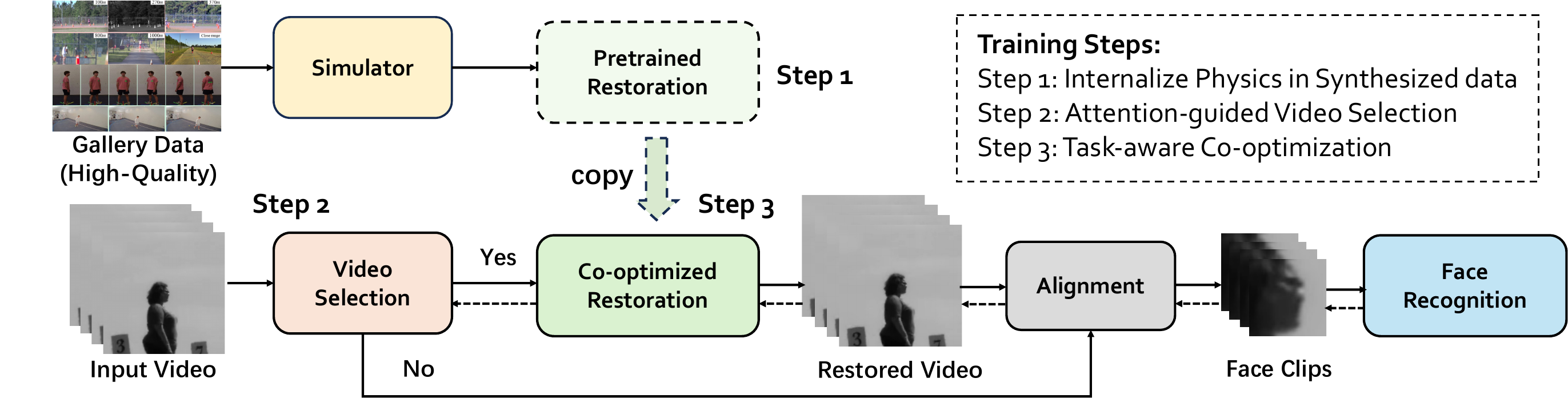}
\caption{\textbf{The overall tri-pillar pipeline} consists of: (i) an optics-accurate turbulence simulator (PATSv5) \cite{roddier1981v,chimitt2024scattering}, (ii) a neuroscience-inspired, attention-guided video selection module \cite{friston2010free,bastos2012canonical}, and (iii) a task-aware co-optimization loop with differentiable alignment \cite{wang2018cosface,deng2019arcface}. \textit{Images shown
with subject permission for publication.}}
\label{fig:overall}
\end{figure*}

Treating restoration and recognition separately may not effectively address the problem. Thus, we propose a tri-pillar framework (Figure~\ref{fig:overall}) and demonstrate the effectiveness of each component in Table~\ref{tab:overall}. In Sections~\ref{sec:signal}, \ref{sec:computer_vision_angel}–\ref{sec:neuroscience}, we introduce each component and answer \textbf{how restoration can be guided to better support recognition from multiple perspectives}.

% Thus, treating restoration and recognition models separately may not effectively solve the problem. Therefore, we propose a tri-pillar framework as shown in Figure~\ref{fig:overall} and illsutrate the effectiveness in Table~\ref{tab:overall}.
% In the following Sections~\ref{sec:signal}, \ref{sec:computer_vision_angel}-\ref{sec:neuroscience}, we explore \textbf{how restoration can be guided to better support recognition from different angles}.

% \begin{table}[h]
%     \centering
%     \caption{Comparison Before and After Using co-optimization strategy, tested on Test Set EVP 4.2.0 Reduced. \textbf{Each column adds on top of the previous configuration.}}
%     \renewcommand{\arraystretch}{1.1}
%     \setlength{\tabcolsep}{2pt}
%     \resizebox{\textwidth}{!}{
%     \begin{tabular}{l | c |c| c|c }
%         \toprule
%         \textbf{Metric} & \textbf{\makecell{Degraded}}& \textbf{\makecell{$+$Restoration}} & \textbf{\makecell{$+$VIDCLS}} & \textbf{\makecell{$+$Co-Optimize}}   \\
%         \midrule
%         \multirow{1}{*}{1:1 TAR@0.1\%FAR$\uparrow$} 
%          & 62.6\% & 63.5\% & 63.3\% & \textbf{64.1\%} \\
%         % \midrule
%         \multirow{1}{*}{1:N Rank Top 20$\uparrow$} 
%          & 87.2\% & 86.5\%  & 86.9\% & \textbf{87.4\%} \\
%         % \midrule
%         \multirow{1}{*}{1:N Search FNIR@1\%FPIR$\downarrow$} 
%          &51.2\% & 51.6\%  & 49.9\% & \textbf{49.9\%} \\
%             \midrule
%          $\Delta$ v.s. Degraded & -- & $-$0.07\% & $+$0.57\% & $+$1.0\%\\
%         \bottomrule
%     \end{tabular}}
%     \label{tab:overall}
% \end{table}

\begin{table}[h]
  \centering
  \caption{Comparison Before and After Using co-optimization strategy, tested on Test Set EVP 4.2.0 Reduced. \textbf{Each column adds on top of the previous configuration.}}
  \renewcommand{\arraystretch}{1.1}
  \setlength{\tabcolsep}{2pt}
  % @{\extracolsep{\fill}} 会把各列平铺到总宽度
    \resizebox{1.\linewidth}{!}{
    \begin{tabular}{l | c |c| c|c }
        \toprule
        \textbf{Metric} & \textbf{\makecell{Degraded}}& \textbf{\makecell{$+$Restoration}} & \textbf{\makecell{$+$VIDCLS}} & \textbf{\makecell{$+$Co-Optimize}}   \\
        \midrule
        \multirow{1}{*}{1:1 TAR@0.1\%FAR$\uparrow$} 
         & 62.6\% & 63.5\% & 63.3\% & \textbf{64.1\%} \\
        % \midrule
        \multirow{1}{*}{1:N Rank Top 20$\uparrow$} 
         & 87.2\% & 86.5\%  & 86.9\% & \textbf{87.4\%} \\
        % \midrule
        \multirow{1}{*}{1:N Search FNIR@1\%FPIR$\downarrow$} 
         &51.2\% & 51.6\%  & 49.9\% & \textbf{49.9\%} \\
            \midrule
         $\Delta$ v.s. Degraded & -- & $-$0.07\% & $+$0.57\% & $+$1.0\%\\
        \bottomrule
    \end{tabular}}
  \label{tab:overall}
\end{table}

\section{Signal Processing Angle: Degradation Representation and In-Loop Modeling}\label{sec:signal}

From a signal processing perspective, explicit modeling of degradation in the signal formation process offers a principled and interpretable approach to improve the robustness of the overall system — encompassing both restoration and recognition components. By integrating domain knowledge into the learning framework, we are motivated to ask the following two key research questions:

\begin{tcolorbox}[before skip=2mm, after skip=0.0cm, boxsep=0.0cm, middle=0.0cm, top=0.1cm, bottom=0.1cm]
    \textit{\textbf{(Q1)}
    If we have accurate analytical modeling of degradations, can it benefit restoration and recognition?
    }\\
    \textit{\textbf{(Q2)}
    What is the best way to utilize analytical modeling? Is it necessary to have it explicitly in the loop?
    }
\end{tcolorbox}
% \vspace*{0.2cm}

\subsection{Model of Turbulence Imaging}
Turbulence manifests in observations taken over a long distance in the form of a spatio-temporally varying blur, acting as a linear operation. This blur is often decomposed into two constituent components: a spatially varying \textit{tilt} (i.e., a pixel shift) and spatially varying blur kernel. These two components give turbulent sequences of images a characteristic ``mirage-like'' quality. Due to Kolmogorov \cite{Kolmogorov_1941_a, Kolmogorov_1941_b}, turbulence is often described as a Gaussian random process and, accordingly, as are waves propagated through turbulence \cite{Tatarski_1967_a}. We now restrict ourselves to a single frame $\vx_f$ in the sequence $\vx$ for notational simplicity.

The spatial varying blur, also known as a point spread function (PSF), is related to the incident wave by a squared Fourier magnitude. Although highly sophisticated models can describe turbulence effects on the amplitude and phase, in weak to moderate turbulence only the phase of the wave is primarily affected. It is then common to write the phase as a sum of Zernike polynomials \cite{noll1976zernike}, basis polynomials that are orthogonal over the often circular pupil of the imaging system. Taking into account the spatio-temporal variation of the distortion, for a given $\vrho = (\rho_h, \rho_w)$ indexing the appropriate vertical and horizontal position in $\vx_f$, the phase may be written as
\begin{equation}
    \vphi_{\vrho} = \sum_{\ell=1}^L \; \underbrace{a_{\vrho, \ell}}_{\text{Per-pixel}} \; \times \; \underbrace{\mZ_{\ell}}_{\text{Fixed}},
    \label{eq: spat_var_phase}
\end{equation}
where $\mZ_{\ell}$ is the $\ell$th Zernike polynomial defined over a grid that is related to the camera parameters (separate from the pixel height and width of the image). This representation can be understood as a basis coefficient \textit{vector} per-pixel $\vp$.

\begin{figure}[t]
\centering
\includegraphics[width=1.\linewidth]{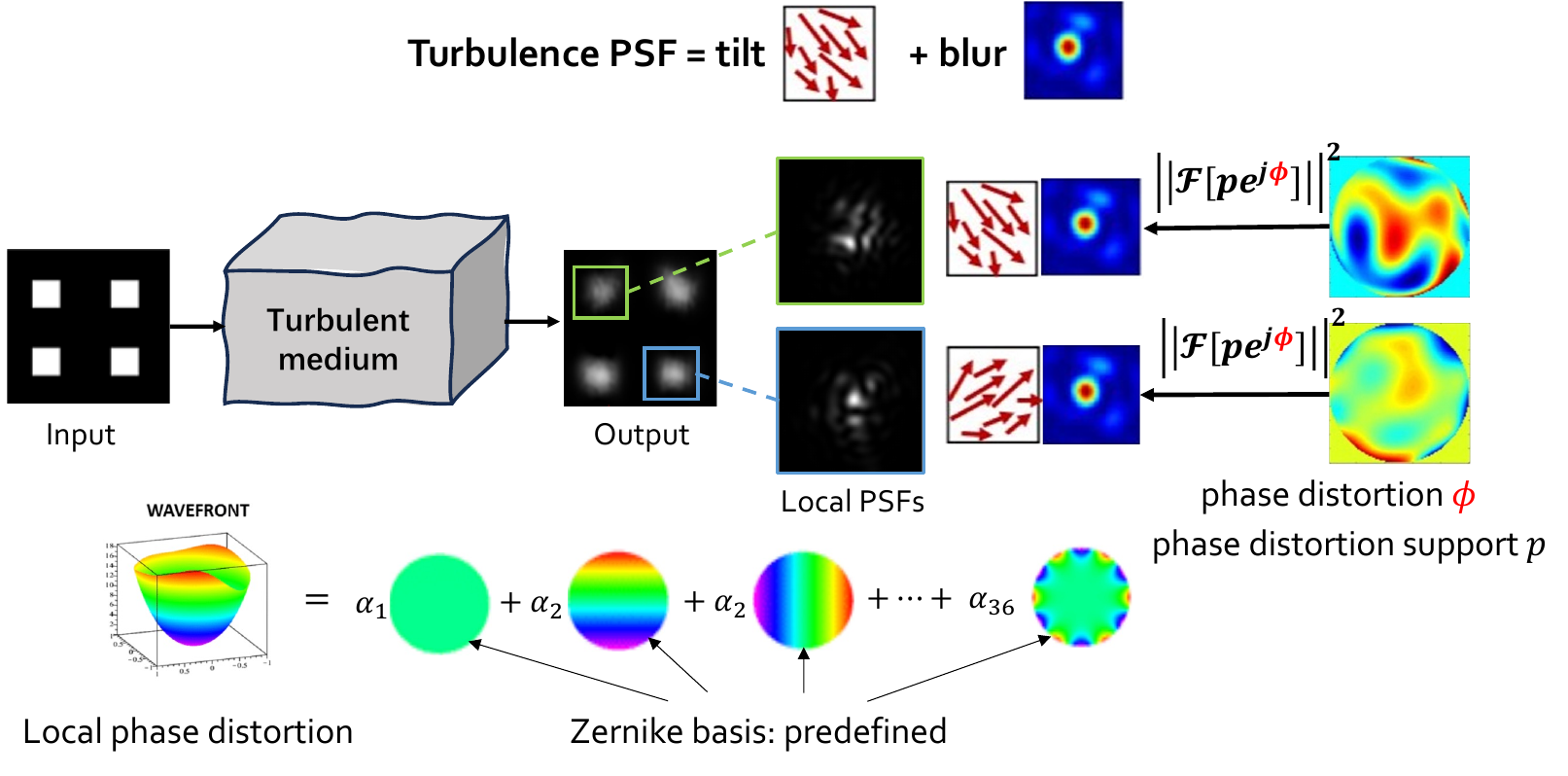}
\vspace{-5mm}
\caption{\textbf{Principle of atmospheric turbulence simulation}~\cite{chimitt2020simulating}. In optics perspective the PSF is caused by distortion on the phase plane, and phase distortion can be represented by linear combination of Zernike basis.}
\label{fig: simulator}
\end{figure}

Our simulator, the Purdue Atmospheric Turbulence Simulator (PATS)~\cite{chimitt2020simulating, chan_turbulence, Mao_2021_a, chimitt2022real}, is a Zernike-based simulation framework, as illustrated in \fref{fig: simulator} that follows the spatially varying phase decomposition \eqref{eq: spat_var_phase}. In this context, the state of the turbulent degradation can be described by sampling $a_{\vrho,\ell}$ $\forall \vrho,\ell$ according to the known statistics \cite{chan_turbulence}, represented as vector $\va \in \R^{H \times W \times L}$. The sampled Zernike space $\va_f$ for a frame $f$ only \textit{parameterizes} the spatially varying blur, hence we must convert $\va$ to an image-space representation. To do this, the PATS model similarly decomposes the spatially varying PSFs using basis kernels $\vpsi_k$ and coefficients $\beta_{\vrho, k}$ similar to \eqref{eq: spat_var_phase}. The collection of PSF basis coefficients is represented as a vector $\vbeta_f \in \R^{H \times W \times K}$ where $K$ denotes the number of basis kernels required to accurately represent turbulent PSFs.

The simulator is described as follows. The simulator takes as input a ground-truth video frame $\mathbf{x}_f \in \mathbb{R}^{H \times W}$ at index $f$, and samples the Zernike space $\va_f$ according to imaging and camera geometries. The Zernike space is then mapped as $\va_f \to (\vtau_f, \vbeta_f)$ following the tilt and blur decomposition with $\vtau_f \in \R^{H \times W \times 2}$ representing the local pixel shift (the two channels as horizontal and vertical shifts). The mappings and $\va_f \to \vtau_f$ and $\va_f \to \vbeta_f$ may be done independently with the former according to camera parameters and the latter using a neural mapping as in \cite{Mao_2021_a}. With $(\vtau_f, \vbeta_f)$ the representation is made to be in image-space and the turbulence-degraded observation $\mathbf{y}_f$ is expressed as \cite{chimitt2022real, chimitt2024scattering}:
% \begin{equation}
%     \mathbf{y}_f = \sum_{k=1}^{K} \; \underbrace{\vpsi_k}_{\text{PSF basis kernel}} \circledast \left( \underbrace{\vbeta_{f,k}}_{\text{Basis coefficient field}} \odot \underbrace{\warp(\mathbf{x}_f; \vtau_f)}_{\text{Geometrically distorted image}}\right) + \mathbf{e},
%     \label{eq: p2s_psf_basis}
% \end{equation}
\begin{equation}
\scalebox{0.9}{$
    \mathbf{y}_f = \sum_{k=1}^{K} \; 
    \underbrace{\vpsi_k}_{\text{PSF basis kernel}} 
    \circledast \left( \underbrace{\vbeta_{f,k}}_{\text{Basis coefficient field}} 
    \odot \underbrace{\warp(\mathbf{x}_f; \vtau_f)}_{\text{Geometrically distorted image}}\right) 
    + \mathbf{e},
$}
\label{eq: p2s_psf_basis}
\end{equation}
where $\warp(\cdot;\vtau_f)$ is the spatial warp operation guided by the pixel-shift field $\vtau_f$, $\vbeta_{f,k} \in \R^{H \times W}$ is the $k$th channel of the vector $\vbeta_f$, $\circledast$ denotes channel-wise convolution, $\odot$ is an element-wise product, and $\mathbf{e}$ is additive white noise. The primarily role of \eqref{eq: p2s_psf_basis} is to approximate the spatially varying ``convolution'' as a sum of invariant convolutions, making the simulator highly efficient ($1000 \times$ or more faster than full wave-based simulations \cite{zhang2025learning}). 

\subsection{Internalizing Degradation Modeling by Generating Synthetic Data}
With PATS, we can construct large-scale datasets to train restoration neural networks for mitigating turbulence effects. We propose a systematic synthetic data generation strategy~\cite{jaiswal2023physics} that captures a wide range of camera parameters and atmospheric conditions, represented by measurable indicators such as $r_0$. By leveraging synthetic datasets that closely match real-world turbulence, we can design specialized modules to better handle these degradations. In this way, the characteristics of turbulence can be internalized by restoration networks. 

To this end, the key factor in our successful improvement of face recognition was the incorporation of accurate physics priors in our data synthesis process, enabling the supervised restoration model to achieve better generalization capability. To validate this approach, we conducted experiments training the restoration model (GRTM network) on two different synthetic datasets: turbulence mitigation transformer (TMT)'s dynamic scene dataset \cite{zhang_2024_TMT} and the ATSyn-dynamic dataset \cite{zhang2024spatio}. Both datasets utilized identical source videos from \cite{SVW} with synthesized turbulence degradation, \textit{differing only in the simulator version employed}. For TMT's dataset, we implemented the PATSv3 simulator \cite{chimitt2022real}, while the ATSyn dataset was generated using PATSv5, 
which incorporates the scattering model of spatially varying blur synthesis~\cite{chimitt2023continuousCn2} and more accurate turbulence path modeling~\cite{chimitt2024scattering}.
% which incorporates a more accurate $r_0$ path \cite{chimitt2023continuousCn2} and scattering model \cite{chimitt2024scattering} for spatially varying blur synthesis. 
This enhanced physics modeling in PATSv5 provides a \textbf{stronger physics prior} compared to PATSv3. As demonstrated in Table \ref{tab:different_simulator}, our experimental results confirm that improved physics priors robustly enhance recognition performance.

% \begin{table}[h]
% \centering
% \caption{
% Face recognition is evaluated on degraded data using restoration models trained with (1) explicit loop modeling and (2) internalized synthetic data. Two different turbulence simulators, PATSv3 and PATSv5, are employed. To minimize confounding factors introduced by non-turbulence content, the experiment is conducted exclusively on a curated subset of long-range turbulence videos.
% }
% \resizebox{0.49\textwidth}{!}{
% \renewcommand{\arraystretch}{1.2}
% \setlength{\tabcolsep}{2pt}
% \begin{tabular}{l|c|cc|cc}
% \toprule
% \multirow{2}{*}{\textbf{\makecell{Face \\Retrieval}}} & \multirow{2}{*}{\textbf{Degraded}} & \multicolumn{2}{c|}{\textbf{Explicit Modeling}} & \multicolumn{2}{c}{\textbf{Internalizing}} \\
%    & & PATSv3 [45] & PATSv5 [70] & PATSv3 [45] & PATSv5 [70] \\
% \midrule
% Rank 5 & 37.75\%  & 39.01\% & 39.19\% & 38.83\% & 39.18\%\\
% Rank 10 & 40.59\% & 41.99\% & 42.16\%  & 41.83\% & 42.18\%\\
% Rank 20 & 45.29\%  & 46.79\% & 46.68\% & 46.40\% & 46.70\%\\
% \bottomrule
% \end{tabular}}
% \label{tab:different_simulator}
% \end{table}

\begin{table}[h]
\centering
\caption{
Face recognition is evaluated on degraded data using restoration models trained with (1) internalized synthetic data and (2) explicit loop modeling. Two different turbulence simulators, PATSv3 and PATSv5, are employed. To minimize confounding factors introduced by non-turbulence content, the experiment is conducted exclusively on a curated subset of long-range turbulence videos.
}
\resizebox{0.9\textwidth}{!}{
\renewcommand{\arraystretch}{1.2}
\setlength{\tabcolsep}{2pt}
\begin{tabular}{l|c|cc|cc}
\toprule
\multirow{2}{*}{\textbf{\makecell{Face \\Retrieval}}} & \multirow{2}{*}{\textbf{Degraded}} & \multicolumn{2}{c|}{\textbf{Internalizing}} & \multicolumn{2}{c}{\textbf{Explicit Modeling}} \\
   & & PATSv3 [45] & PATSv5 [70] & PATSv3 [45] & PATSv5 [70] \\
\midrule
Rank 5 & 37.75\%   & 38.83\% & 39.18\%& 39.01\% & 39.19\%\\
Rank 10 & 40.59\%   & 41.83\% & 42.18\%& 41.99\% & 42.16\%\\
Rank 20 & 45.29\%   & 46.40\% & 46.70\%& 46.79\% & 46.68\%\\
\bottomrule
\end{tabular}}
\label{tab:different_simulator}
\end{table}

\subsection{Explicit Degradation Modeling in the Loop}

\begin{figure}[h]
\centering
\includegraphics[width=.85\linewidth]{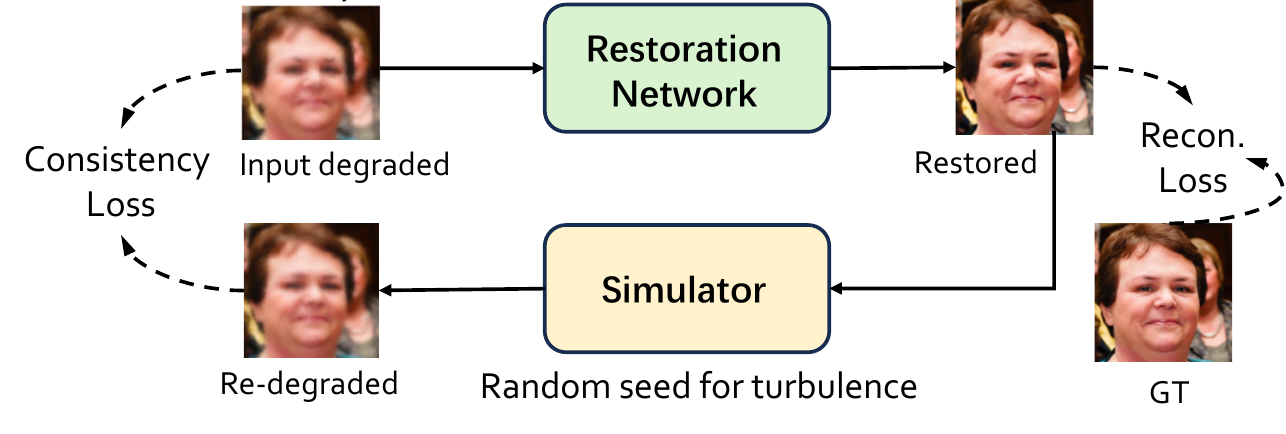}
\caption{The principle of \textbf{explicitly modeling degradation in the restore-degrade loop}~\cite{jaiswal2023physics}, where a physics-based simulator re-applies degradation to the restored output to compute consistency loss. \textit{Images shown
with subject permission for publication.}}
% \vspace{-3mm}
\label{fig: simulator_in_loop}
\end{figure}

To effectively mitigate real-world degradations, our framework further tries to explicitly integrate physics-based simulators into the restoration pipeline \cite{jaiswal2023physics}. Specifically, in the training stage as shown in Figure~\ref{fig: simulator_in_loop}, we re-degrade the reconstructed image by the differentiable simulator and align it with the original input to enforce the consistency between image formation and restoration. Our consistency enforcement could facilitate the separation of image semantics and turbulence profiles by injecting the turbulence conditions into the training loop. During training, the consistency error can be back-propagated through the degradation model, carrying the gradient with physics property provided by the simulator to the restoration model to help it learn from the physics. This physics-informed approach creates a closed feedback loop that enhances the model's understanding of the underlying degradation mechanisms. With the physics-integrated restoration network, experimental results illustrated in \cite{jaiswal2023physics} that the above strategy significantly improves the generalization across multiple real-world datasets with varying turbulence strength. The incorporation of physical priors enables our model to better handle the complex and stochastic nature of atmospheric turbulence, resulting in more robust performance on previously unseen degradation conditions.
\revise{As demonstrated in Table~\ref{tab:different_simulator}, explicit modeling exhibits a distinct advantage over the internalized approach when using the less sophisticated PATSv3 simulator (e.g., reaching 39.01\% at Rank 5 versus 38.83\%). This suggests that the consistency loop more effectively leverages even imperfect physics-based gradients to guide the restoration. However, this advantage diminishes as the simulator fidelity improves; under the more advanced PATSv5, both methods achieve comparable performance (39.19\% vs 39.18\% at Rank 5). While explicit modeling offers a stronger inductive bias for handling complex degradations, it also introduces higher computational overhead during training due to the back-propagation through the differentiable simulator. Consequently, while explicit modeling is more robust to weaker physical priors, its marginal utility decreases when high-fidelity synthetic data is available.}

% As shown in Table~\ref{tab:different_simulator}, explicit modeling improves recognition performance and becomes more effective as the accuracy of physical priors increases. Moreover, compared to the internalized modeling approach, explicit modeling yields greater improvements under PATSv3 (a weaker physical prior), while achieving comparable performance under PATSv5 (a stronger physical prior).

\begin{tcolorbox}[before skip=2mm, after skip=0.0cm, boxsep=0.0cm, middle=0.0cm, top=0.1cm, bottom=0.1cm]
    \textit{\textbf{Takeaways:} \\
    % \begin{itemize}[leftmargin=*]
    \textit{\textbf{(A1)}} \textbf{Precise signal modeling} leads to improved performance in both restoration and recognition.\\
        % \item \textbf{Precise signal modeling} is very helpful for both restoration and recognition goals.
       \textit{\textbf{(A2)}}
       Both \textbf{explicit and internalizing} approaches are effective; when physical modeling is sufficiently precise, internalizing methods can achieve comparable performance by learning from its outputs.
    % \end{itemize}
    }
\end{tcolorbox}

% \begin{tcolorbox}[before skip=2mm, after skip=0.0cm, boxsep=0.0cm, middle=0.0cm, top=0.1cm, bottom=0.1cm]
%     \textit{\textbf{(A1)}
%     \textbf{More precise signal modeling} can lead to improved performance in both restoration and recognition tasks.
%     }\\
%     \textit{\textbf{(A2)}
%     Both \textbf{explicit and internalizing} ways can work; internalizing can match performance when physical modeling is precise enough.
%     }
% \end{tcolorbox}

\section{Computer Vision Angle: Crossing the Semantic Gap}
\label{sec:computer_vision_angel}
% Moreover, real-world degradations often exhibit diverse types and varying severities, which are difficult to capture through explicit modeling alone.

% While restoration and recognition are often treated separately, we observe that a naive combination — such as pre-training a restoration model followed by recognition fine-tuning — can lead to \textit{catastrophic forgetting} of low-level knowledge that is essential for handling real-world degradations. This raises the question of how to jointly optimize both tasks without sacrificing one for the other. In this section, we explore how multi-task learning and scalable training with synthetic and real-world data can enable restoration knowledge to be preserved and effectively transferred across the semantic gap to improve high-level recognition.
Physical degradation modeling in signal formation can enhance robustness to some extent.
However, there remains a risk of mismatch between real-world data and the assumed physical model, which can limit its effectiveness.
To address this, we approach the restoration–recognition problem from a computer vision perspective. By leveraging co-optimization on both synthetic and real-world data, we aim to bridge the semantic gap. This allows restoration knowledge to be preserved and effectively transferred to improve high-level recognition performance.
Thus, these considerations motivate the following three key questions:

\begin{tcolorbox}[before skip=2mm, after skip=0.0cm, boxsep=0.0cm, middle=0.0cm, top=0.1cm, bottom=0.1cm]
    \textit{\textbf{(Q3)}
    Is data augmentation alone sufficient to train a good recognition model?
    }\\
    \textit{\textbf{(Q4)}
    Shall we formulate restoration$+$recognition as joint optimization? 
    }\\
    \textit{\textbf{(Q5)}
    How do we leverage both synthetic and real-world data?
    }
\end{tcolorbox}

% \begin{table*}[h]
% \centering
% \caption{Ablation in the training dataset and the augmentation. Comparison between evaluation \textbf{without} and \textbf{with} restoration.}
% \renewcommand{\arraystretch}{1.2}
%     \setlength{\tabcolsep}{2pt}
% \resizebox{.85\textwidth}{!}{
% \begin{tabular}{l|cccc|cccc}
% \toprule
% \textbf{\multirow{2}{*}{Method}}& \multicolumn{4}{c|}{\textbf{w/o Restoration}} & \multicolumn{4}{c}{\textbf{w/ Restoration}} \\
% \cmidrule(lr){2-5} \cmidrule(lr){6-9}
%  & Rank20$\uparrow$ & TAR@FAR0.001$\uparrow$ & TPIR@FNIR0.01$\uparrow$ & Mean$\uparrow$ & Rank20$\uparrow$ & TAR@FAR0.001$\uparrow$ & TPIR@FNIR0.01$\uparrow$ & Mean$\uparrow$\\
% \midrule
% ViT + Augmentation                  & 65.09 & 28.99 & 16.70 & 36.93& 67.23 & 31.13 & 18.08 & 38.81 \\
% ViT + Aug. + LQ Dataset             & 73.90 & 32.58 & 17.70 & 41.39& 74.53 & 32.83 & 17.88 & 41.75\\
% ViT + Aug. + LQ Dataset + Rest.     & 73.21 & 35.22 & 19.16 & 42.53& 73.58 & 35.53 & 19.30 & 42.80 \\
% % \midrule
% % \textbf{Mean}                       & 36.93 & 41.39 & 42.53 & 38.81 & 41.75 & 42.80 \\
% \bottomrule
% \end{tabular}}
% \label{tab:frwithoutevalrestor}
% \end{table*}

\subsection{Scaling Up Training Data}
\noindent\textbf{Is Data Augmentation Alone Sufficient for Recognition?}
We investigate how training strategies and image restoration influence face recognition on low-quality (LQ) images. A Vision Transformer (ViT) model is trained under three settings: (1) baseline with standard augmentations (crop, resize, color jitter, contrast), (2) adding a LQ dataset from BRIAR to the WebFace4M~\cite{zhu2022webface260m} base, and (3) further applying restoration as an augmentation to the LQ data. Each setup is evaluated with and without test-time restoration (Table~\ref{tab:frwithoutevalrestor}). Without restoration, adding the LQ dataset improves mean performance from 36.93 to 41.39, and Rank-20 from 65.09 to 73.90. Restoration-based augmentation raises the mean to 42.53 and TAR@FAR=0.001 to 35.22. Applying restoration at test time yields further gains, with the full setup achieving a mean of 42.80 and TAR@FAR=0.001 of 35.53. These findings underscore that integrating a tailored LQ dataset and restoration — both during training and at test time — optimizes face recognition performance, offering a robust solution for real-world degraded image scenarios.

\vspace{1mm}
\noindent\textbf{Combining Synthetic Data and Real Data.}
The results suggest that while data augmentation plays a crucial role in improving face recognition performance under degraded conditions, it is not sufficient on its own. The addition of a LQ dataset significantly enhances recognition accuracy, highlighting the importance of training models on diverse, real-world degraded data. Furthermore, incorporating restoration as an augmentation technique provides an additional boost, particularly when test-time restoration is applied. This indicates that a holistic approach-combining augmentation, LQ dataset training, and restoration-yields the most robust recognition performance. However, the improvements from restoration are relatively modest, suggesting that while it helps, its effectiveness depends on the degree of degradation and the alignment between restoration and recognition objectives.
These findings highlight the need for task-specific restoration that preserves identity-critical features rather than improving perceptual quality alone.

% These findings reinforce the need for task-specific restoration strategies that preserve identity-critical features rather than purely enhancing image quality based on traditional perceptual metrics. 

\subsection{Co-Optimization on Restoration and Recognition}

\begin{figure}[h]
\centering
\includegraphics[width=1.\linewidth]{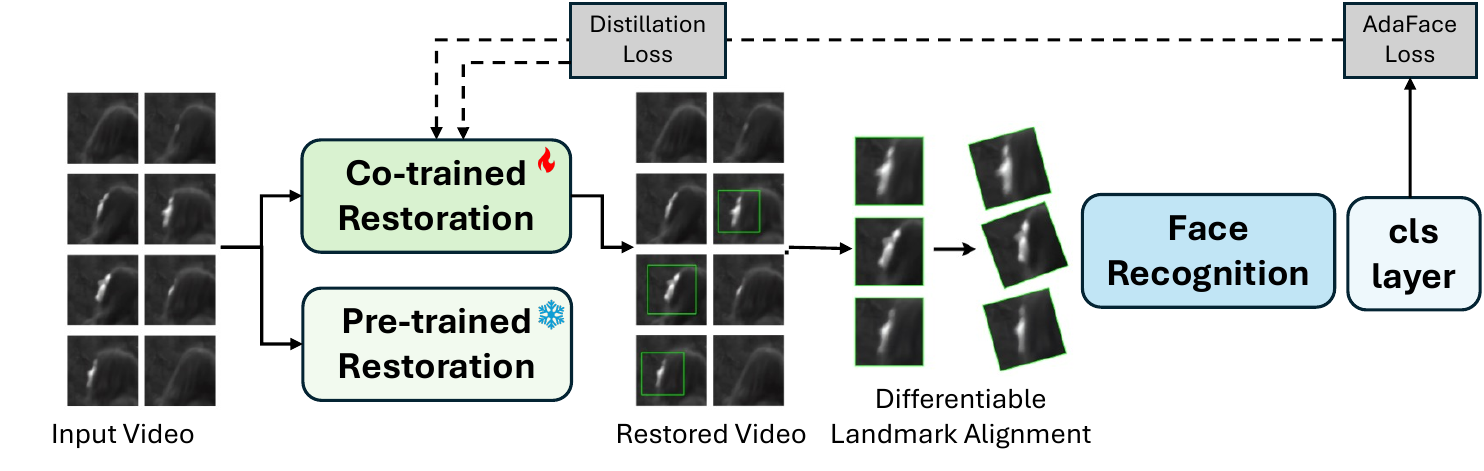}
\caption{\textbf{Co-optimization pipeline for restoration and recognition.} The input video undergoes either co-trained restoration or pre-trained restoration, with distillation loss supervising the co-trained model. The restored video is processed through differentiable landmark alignment before face recognition, which is optimized using AdaFace loss at the classification layer. Not all frames may have a face detection, and such frames are filtered out before proceeding to the face recognition module. \textit{Images shown
with subject permission for publication.}}
\label{fig:co-training}
\end{figure}

\noindent\textbf{Overall Co-Optimization Framework.}
Unlike traditional pipelines that treat restoration and recognition as sequential and independent tasks, our framework integrates these processes into a unified pipeline. By co-optimizing restoration and recognition, we ensure that restoration efforts are directly aligned with downstream recognition objectives as shown in Figure~\ref{fig:co-training}. This synergy is achieved through targeted loss functions and adaptive mechanisms that enable dynamic improvement based on task-specific requirements.
The restoration module in our framework is trained with a dual loss function:
\begin{itemize}
    \item \textbf{Perceptual Loss ($\mathcal{L}_{\text{percep}}$)}: Ensures that restored images meet visual quality standards.
\item \textbf{Recognition Loss ($\mathcal{L}_{\text{adaface}}$)}: Aligns restoration outputs with features essential for downstream tasks, such as identity-preserving details in face recognition.
\end{itemize}
Specifically, the degraded video frames $\vy$ are first processed by the restoration model. The restored image is then aligned using facial landmarks and forwarded to the face recognition module, where a biometric feature is extracted and the biometric loss $\mathcal{L}_{\text{adaface}}$ is computed.
The overall co-training loss is defined as:
\begin{equation}
\mathcal{L}_{\text{total}} = \mathcal{L}_{\text{adaface}} + \lambda\mathcal{L}_{\text{percep}} ,
\label{eq:whole_loss}
\end{equation}

\begin{table}[h]
    \centering
    \caption{Comparison among separately pretrained restoration, fine-tuning the restoration model using recognition loss only, co-optimization with a restore-first strategy, and co-optimization with an align-first strategy, evaluating both recognition performance and visual quality (PSNR$\uparrow$ and SSIM$\uparrow$).}
    \renewcommand{\arraystretch}{1.2}
    \setlength{\tabcolsep}{2pt}
    \resizebox{0.8\textwidth}{!}{
    \begin{tabular}{l | c |c| c|c}
        \toprule
        \textbf{Metric} & \textbf{\makecell{Pretrained}} & \textbf{\makecell{Fine-tune}} & \textbf{\makecell{Co-optimize \\ Restore-first}} &  \textbf{\makecell{Co-optimize\\ Align-first}} \\
        \midrule
        \multirow{1}{*}{1:1 TAR@0.1\%FAR$\uparrow$} 
         & 63.5\% & 63.6\% & \textbf{64.1\%} & 63.7\% \\
        % \midrule
        \multirow{1}{*}{1:N Rank Top 20$\uparrow$} 
         & 86.5\%  & 86.8\%& \textbf{87.4\%} & 86.5\%\\
        % \midrule
        \multirow{1}{*}{1:N Search FNIR@1\%FPIR$\downarrow$} 
         & 51.6\%  & 52.5\% & \textbf{49.9\%}  & 51.4\% \\
                 \midrule
        PSNR$\uparrow$  & 31.94& 15.69 & \textbf{32.00} & 31.59\\
         SSIM$\uparrow$  &  \textbf{0.8799} & 0.2386 & 0.8790 & 0.8785 \\
        \bottomrule
    \end{tabular}}
    \vspace{-2mm}
    \label{tab:cotraining}
\end{table}

\noindent where $\lambda$ is the weighting coefficient that balances the contributions of the two loss terms. The total loss is back-propagated to update the restoration network.
To improve consistency, a differentiable landmark alignment step is also integrated into the training pipeline, ensuring that faces are properly rotated before being passed to the recognition model.

% We also explore various settings on co-training experiments, which will be explained in details in following sections and illustrated in Figure~\ref{fig:settings}.

\vspace{1mm}
\noindent\textbf{Self-Distillation Framework.}
In practice, we aim to leverage real-world degradation data for co-training to mitigate domain gap issues. However, when such real data lacks paired degraded/clean images, traditional $\ell_1$ or $\ell_2$ reconstruction losses cannot be directly applied between the restored output and a clean ground truth.
If we simply remove the perceptual loss $\mathcal{L}_{\text{percep}}$ in Eq.~(\ref{eq:whole_loss}) and fine-tune the restoration module using only the recognition loss, the restoration ability deteriorates drastically, resulting in \textbf{catastrophic forgetting}.
As shown in Table~\ref{tab:cotraining}, fine-tuning with recognition loss alone (denoted as Fine-tune) leads to poor visual quality and can even degrade overall recognition performance compared to training the restoration model separately on synthetic data (denoted as Pretrained), simulated using the method described in Section~\ref{sec:signal}. This is likely due to the limited diversity and quantity of the co-training data, which causes the model to overfit to the insufficient fine-tuning set.

To address this challenge, we propose a self-distillation framework that employs Siamese twin models to compute the distillation loss, serving as the \textbf{perceptual loss} within the overall co-training architecture. 
This loss imposes constraints on the perceptual quality of the restored results, ensuring that the co-trained restoration model can still produce visually pleasing outputs.
Specifically, we create two instances of our pre-trained video restoration model: one remains frozen as the teacher model $\mathcal{R}_{\theta_{\text{teacher}}}$, while the other is fine-tuned for co-optimization as the student model $\mathcal{R}_{\theta_{\text{student}}}$. The pixel-level loss is then computed as the $l_1$ distance between their outputs, serving to preserve the model’s restoration capability as following:
\begin{equation}
    \mathcal{L}_{\text{percep}} = \|\mathcal{R}_{\theta_{\text{teacher}}}(\vy) - \mathcal{R}_{\theta_{\text{student}}}(\vy)\|_1.
\end{equation}
\revise{To this end, the restoration ability is distilled from the pre-trained teacher model to the co-trained student model, ensuring that \textit{the student model maintains the visual quality of the restored results while improving its recognition performance}, denoted as Co-optimize in Table~\ref{tab:cotraining}.}

% \begin{table*}[!h]
% \centering
% \caption{Ablation on loss weight $\lambda$ during Phase II and Phase III. Bold numbers indicate the best performance in each metric.}
% \label{tab:ablation_loss_weight}
% \begin{adjustbox}{width=\linewidth}
% \begin{tabular}{l|cccc}
% \toprule
% \textbf{$\lambda$} & 
% \textbf{Phase II: TAR@FAR=0.1\%} & 
% \textbf{Phase II: FNIR@TPIR=1.0\%} & 
% \textbf{Phase III: TAR@FAR=0.01\%} & 
% \textbf{Phase III: FNIR@TPIR=0.3\%} \\
% & \textbf{(fusion) (↑)} & \textbf{(fusion) (↓)} & \textbf{(fusion) (↑)} & \textbf{(face-only) (↓)} \\
% \midrule
% 600 (Best for Phase II)   & \textbf{80.9\%} & \textbf{45.1\%} & 60.9\% & 55.4\% \\
% 800                        & 80.8\% & 44.4\% & 60.9\% & 55.3\% \\
% 1,000 (Best for Phase III)& 80.6\% & 44.4\% & \textbf{61.5\%} & \textbf{54.8\%} \\
% \bottomrule
% \end{tabular}
% \end{adjustbox}
% \end{table*}

\begin{table}[h]
\centering
\caption{Ablation on training data and augmentation. Evaluation \textbf{without} vs \textbf{with} restoration. R20, TAR, and TPIR denote Rank-20, TAR@FAR=0.001, and TPIR@FNIR=0.01, respectively.}
\renewcommand{\arraystretch}{1.0}
\setlength{\tabcolsep}{3pt}
% \scriptsize
\resizebox{.9\linewidth}{!}{
\begin{tabular*}{\textwidth}{l|ccc|ccc}
\toprule
\textbf{Method} & \multicolumn{3}{c|}{\textbf{w/o Restoration}} & \multicolumn{3}{c}{\textbf{w/ Restoration}} \\
\cmidrule(lr){2-4} \cmidrule(lr){5-7}
 & R20$\uparrow$ & TAR$\uparrow$ & TPIR$\uparrow$ & R20$\uparrow$ & TAR$\uparrow$ & TPIR$\uparrow$ \\
\midrule
ViT + Aug. & 65.09 & 28.99 & 16.70 & 67.23 & 31.13 & 18.08 \\
+ LQ Data  & \textbf{73.90} & 32.58 & 17.70 & \textbf{74.53} & 32.83 & 17.88 \\
+ LQ + Restoration   & 73.21 & \textbf{35.22} & \textbf{19.16} & 73.58 & \textbf{35.53} & \textbf{19.30} \\
\bottomrule
\end{tabular*}}
\label{tab:frwithoutevalrestor}
\end{table}

\vspace{1mm}
\noindent\textbf{Differentiable Face Alignment.}
We find that another critical component for unlocking restoration-recognition co-training is differentiable face landmark alignment. Face landmark alignment is the process of aligning face landmarks (eyes, nose, and edge of lips) from an original image to fixed coordinates on a grid. This is a highly effective regularization step and omitting it hurts performance. The steps for face landmark alignment are 1) infer the landmarks with a predictor, 2) estimate an affine transformation to transform the predicted landmarks to fixed coordinates, and 3) transform the image onto a $112\times 112$ grid with the estimated affine transformation.

In the context of restoration-recognition co-training, we consider two pipeline configurations for including face alignment: \textit{align-first} and \textit{restore-first}. Align-first is a simpler implementation for co-training because the face loss $\mathcal{L}_{\text{adaface}}$ does not back-propagate through alignment. However, we find that \textit{align-first} has multiple downsides. 
The \textit{align-first} pipeline passes a warped image to the restoration model, where the cropped regions lack surrounding context, thereby impairing the model’s restoration ability.
% The \textit{align-first} pipeline will pass a warped image to the restoration model, where as the pretrained video restoration model ($\vtheta_{\text{teacher}}$ above) was not trained with warped images.
% This is a distribution shift that makes the distillation loss less effective.
Additionally, the \textit{align-first} strategy introduces alignment inconsistencies in videos. If each frame is aligned independently, spatial misalignment occurs across frames. Alternatively, if a single transformation is applied to the entire video, all frames except the reference one will be misaligned.
In contrast, the \textit{restore-first} strategy preserves richer contextual information for restoration and avoids mismatch issues across frames.

% more flexible preprocessing that is consistent with the input distribution used during the pre-training of the video restoration model. Each frame can then be aligned independently before being passed to the face embedding module.

% Additionally, \textit{align-first} causes alignment inconsistencies for videos. If each frame of a video is aligned independently, then the frames will not be spatially aligned. If the entire video is aligned by one transformation, then all frames except one will be misaligned. In contrast to \textit{align-first}, restore-first allows flexible pre-processing to match the input used in the pre-training of the video restoration model. Then each frame can be aligned independently before being passed to the face embedding model.

To unlock \textit{restore-first} co-training, differentiable face landmark alignment is required, since the gradient needs to flow through the alignment module to effectively update the restoration model.
We adopt the differentiable alignment implementation provided by \cite{kprpe} to enable restore-first co-training. With differentiable alignment, the face loss with respect to the restoration model is
\begin{equation}
\frac{\partial \mathcal{L}_{\text{adaface}}}{\partial \mathcal{R}_{\theta_{\text{student}}}} = 
\frac{\partial \mathcal{L}_{\text{adaface}}}{\partial \mathcal{G}_\varphi(\mathcal{R}_{\theta_{\text{student}}}(\vy))} \cdot 
\frac{\partial\mathcal{G}_\varphi(\mathcal{R}_{\theta_{\text{student}}}(\vy))}{\partial \mathcal{R}_{\theta_{\text{student}}}(\vy)} \cdot 
\frac{\partial \mathcal{R}_{\theta_{\text{student}}}(\vy)}{\partial \mathcal{R}_{\theta_{\text{student}}}}
\end{equation}
% \begin{equation}
% \frac{\partial \mathcal{L}_{\text{adaface}}}{\partial \vtheta_{\text{student}}} = \frac{\partial \mathcal{L}_{\text{adaface}}}{\partial \vphi(\vtheta_{\text{student}}(\vy))} \cdot \frac{\partial \vphi(\vtheta_{\text{student}}(\vy))}{\partial \vtheta_{\text{student}}(\vy)} \cdot \frac{\partial \vtheta_{\text{student}}(\vy)}{\partial \vtheta_{\text{student}}}
% \end{equation}
where $\mathcal{G}_\varphi$ is the face recognition model and $\vy$ is the input video clip. The results in Table~\ref{tab:cotraining} demonstrate the effectiveness of restore-first co-optimization, showing that restore-first outperforms align-first.

\begin{figure}[t]
    % \vspace{-3mm}
    \centering
    \includegraphics[width=.8\linewidth]
    {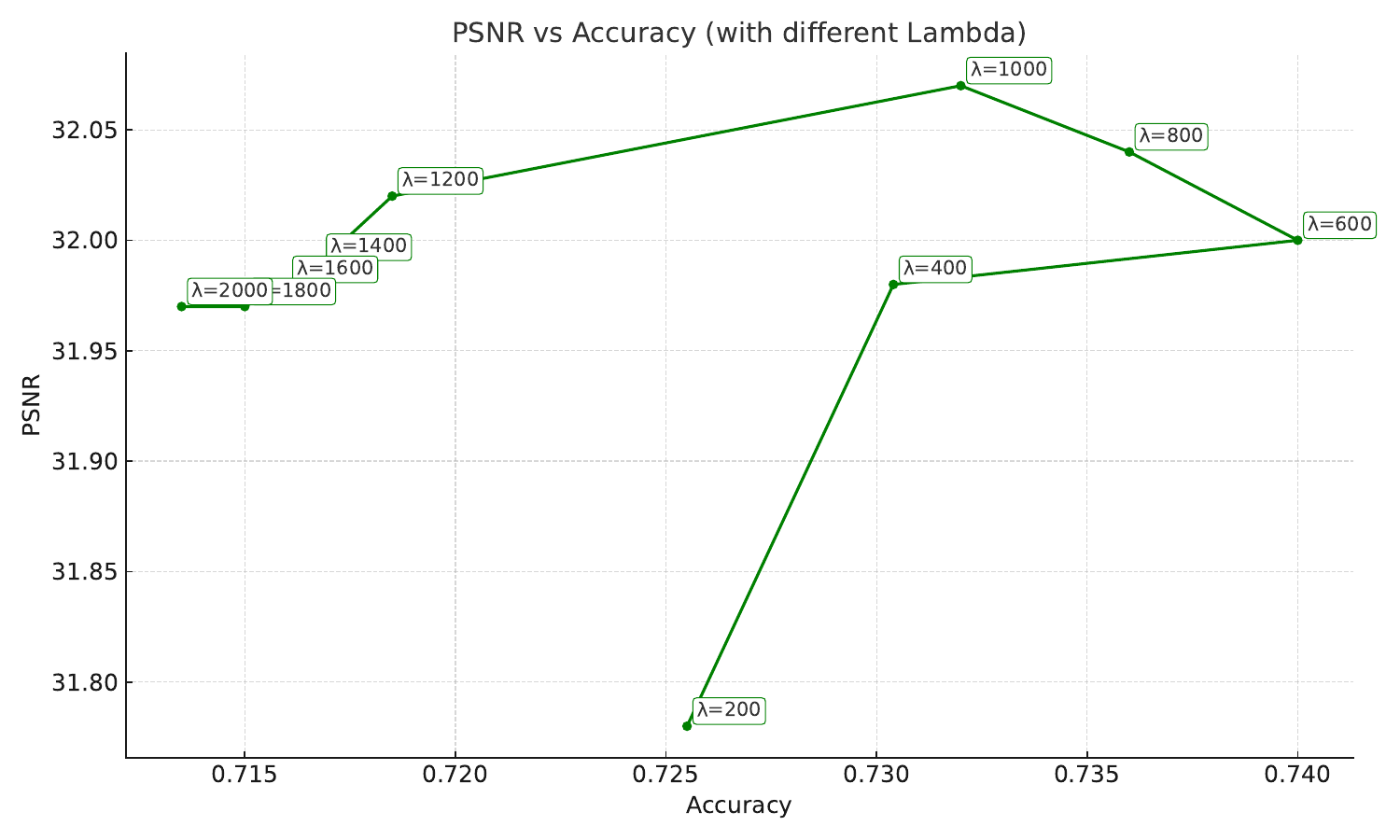}
    % \vspace{-3mm}
    \caption{
     Trade-off between perceptual quality (PSNR) and recognition accuracy as the loss weight $\lambda$ varies. While PSNR remains stable or improves with larger $\lambda$, recognition performance can degrade — highlighting the need to balance restoration and task objectives.
}
    % \vspace{-5mm}
    \label{fig:different_lambda}
\end{figure}

\vspace{1mm}
\noindent\textbf{How to Balance Multiple Tasks?}
To investigate the impact of loss weighting in our co-optimization framework, we conduct ablation studies by varying the relative weight $\lambda$ for restoration and recognition tasks and evaluate performance across a range of $\lambda$ values, as shown in Table~\ref{tab:different_weight} and Figure~\ref{fig:different_lambda}.
First, using a reasonable value of $\lambda$ (to bring the restoration loss and recognition loss to a similar scale) will slightly affect the final results.
These results reveal a fundamental quality-discrimination tradeoff in joint restoration-recognition systems.
As $\lambda$ increases, the visual quality initially improves. However, around $\lambda=1000$, a turning point emerges — further increasing $\lambda$ begins to degrade visual quality. In contrast, recognition accuracy peaks at $\lambda=600$. This is because when $\lambda$ becomes too large, the contribution of the recognition loss diminishes, even though the semantic information from the recognition task can benefit visual quality. Conversely, when $\lambda$ is too small, the perceptual loss provides insufficient supervision, which negatively impacts both recognition and restoration performance. These observations suggest that maintaining a balanced level of restoration capability is essential for supporting downstream tasks effectively.

These findings suggest that the importance of restoration loss increases as precision requirements become more stringent, highlighting the need for task-aware loss balancing that adapts to the target operating regime. Overall, a moderate balance between the two objectives is essential for achieving restoration that is both task-aware and generalizable across varying precision–recall demands.

% \begin{table}[t]
% \centering
% \scriptsize
% \caption{Ablation in the training dataset and the augmentation. Evaluation Images are \textbf{without restoration}. }
% \resizebox{0.48\textwidth}{!}{
% \begin{tabular}{c| c c c}    
% \toprule
%  & \makecell[l]{ViT\\+Augmentation} & \makecell[l]{ViT\\+Augmentation\\+LQ Dataset} & \makecell[l]{ViT\\+Augmentation\\+LQ Dataset\\+Restoration} \\
% \midrule
% Rank20 & 65.09& 73.90& 73.21\\
% TAR@FAR0.001 & 28.99& 32.58& 35.22\\
% TPIR@FNIR0.01 & 16.70& 17.70& 19.16\\\hline
%  MEAN& 36.93& 41.39&42.53\\
% \toprule
% \end{tabular}}
% \label{tab:frwithoutevalrestor}
% \end{table}

% \begin{table}[t]
% \centering
% \scriptsize
% \caption{Ablation in the training dataset and the augmentation. Evaluation Images are \textbf{with restoration}.}
% \resizebox{0.48\textwidth}{!}{
% \begin{tabular}{c| c c c}    
% \toprule
%  & \makecell[l]{ViT\\+Augmentation} & \makecell[l]{ViT\\+Augmentation\\+LQ Dataset} & \makecell[l]{ViT\\+Augmentation\\+LQ Dataset\\+Restoration} \\
% \midrule
% Rank20 &  67.23&  74.53&  73.58
% \\
% TAR@FAR0.001 &  31.13&  32.83&  35.53
% \\
% TPIR@FNIR0.01 &  18.08&  17.88&  19.30
% \\\hline
%  MEAN& 38.81& 41.75&42.80\\
% \bottomrule
% \end{tabular}}
% \label{tab:frwithevalrestor}
% \end{table}

\begin{tcolorbox}[before skip=2mm, after skip=0.0cm, boxsep=0.0cm, middle=0.0cm, top=0.1cm, bottom=0.1cm]
    \textit{\textbf{Takeaways:}\\
    % \begin{itemize}[leftmargin=*]
            \textit{\textbf{(A3)}} \textbf{Data augmentation} helps but remains \textbf{insufficient}; real degraded data and restoration are still needed for recognition. \\
        \textit{\textbf{(A4)}}  \textbf{Jointly optimizing restoration$+$recognition} is necessary. Restore knowledge cannot be forgotten and must be preserved partially for best recognition performance. \\
       \textit{\textbf{(A5)}}  We can scale up joint training even no clean ground-truth for real images – using \textbf{pseudo label suffices to scale up}.
    % \end{itemize}
    }
\end{tcolorbox}

\begin{table}[h]
    \centering
    \caption{Comparison of the co-optimization framework under different weights $\lambda$, evaluating both recognition performance and visual quality (PSNR$\uparrow$ and SSIM$\uparrow$).}
    \renewcommand{\arraystretch}{1.0}
    \setlength{\tabcolsep}{10pt}
    \resizebox{0.9\textwidth}{!}{
    \begin{tabular}{l | c |c| c}
        \toprule
        \textbf{Metric}  & \makecell{$\lambda=600$}& \makecell{$\lambda=800$} & \makecell{$\lambda=1000$}  \\
        \midrule
        \multirow{1}{*}{1:1 TAR@0.01\%FAR$\uparrow$} &  
          60.9\% & 60.9\% & \textbf{61.5\%} \\
        % \midrule
        \multirow{1}{*}{1:N Rank Top 20$\uparrow$} 
        & \textbf{87.4\%} & \textbf{87.4\%} & 87.2\% \\
        % \midrule
        \multirow{1}{*}{1:N Search FNIR@0.3\%FPIR$\downarrow$} 
         & 55.4\% & 55.3\% & \textbf{54.8\%} \\
        \midrule
        PSNR$\uparrow$ & 32.00 & 32.05 &\textbf{32.07}  \\
         SSIM$\uparrow$  & 0.8790& 0.8795& \textbf{0.8803}   \\
    \bottomrule
    \end{tabular}}
    \vspace{-2mm}
    \label{tab:different_weight}
\end{table}

\section{Neuroscience Angle: Attention and Neuroplasticity}\label{sec:neuroscience}

The human visual system exhibits remarkable efficiency and robustness, partly due to its ability to selectively process information \cite{posner1980orienting} and adapt its pathways based on experience and task demands \cite{pascualleone2005plastic}. Drawing inspiration from these neuroscience principles — namely attention and neuroplasticity — can offer valuable insights into designing more effective and efficient restoration-recognition systems \cite{hassabis2017neuroscience, vaswani2017attention}. This perspective challenges the assumptions that all visual input must be exhaustively restored and that the entire restoration model requires adaptation for a downstream task, leading us to another two key questions:

\begin{tcolorbox}[before skip=2mm, after skip=0.0cm, boxsep=0.0cm, middle=0.0cm, top=0.1cm, bottom=0.1cm]
    \textit{\textbf{(Q6)}
     Should we restore all visual inputs for recognition purpose?
    }\\
    \textit{\textbf{(Q7)}
    Should we fine-tune all parameters of the restoration model for better recognition results?
    }
\end{tcolorbox}

\vspace{-2mm}
\subsection{Attention: Bottom Up and Top Down}

\begin{figure}[h]
\centering
\includegraphics[width=0.9\linewidth]{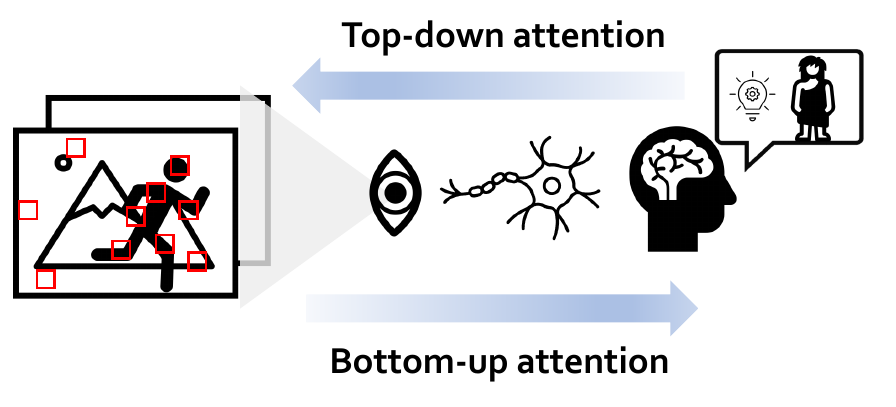}
\caption{\textbf{The human visual system employs attentional selection} to regulate the processing of visual input, assigning greater cognitive resources to stimuli that are salient or relevant to current goals, while attenuating the impact of irrelevant or distracting signals.}
\label{fig:attention}
\end{figure}

Human vision does not process every photon hitting the retina with equal priority. Instead, attentional mechanisms filter incoming sensory data, prioritizing salient or task-relevant information while suppressing distractions \cite{bundesen1990theory, posner1980orienting}. This selective processing is crucial for navigating complex environments efficiently (Figure \ref{fig:attention}). Our restoration-recognition framework integrates analogous selection mechanisms across frames, spatial regions, and entire video sequences.

\vspace{1mm}
\subsubsection{Bottom-up Attention: Frame and Region Selection}
\begin{figure}[t]
\centering
\includegraphics[width=.8\linewidth]{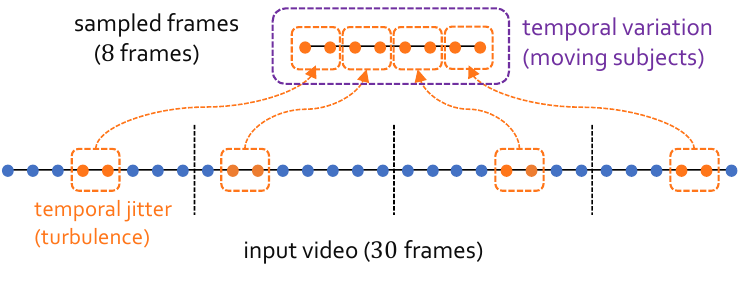}
\caption{Illustration of the \textbf{temporal sampling strategy}, which significantly reduces computational cost by selecting only a subset of representative frames, while preserving the original motion information.}
\label{fig:frame_sample}
\end{figure}

Bottom-up attention is driven by the inherent properties of the stimulus, such as contrast, motion, or uniqueness \cite{itti1998model}. This principle can be applied not only within images but also across time in video sequences. In the context of video restoration for recognition:

\vspace{1mm}
\noindent \textbf{Temporal Frame Selection:} 
% Since the motion in our videos is relatively small, but capturing more informative content under degraded conditions is important, the data collected in BRIAR project is in the form of video frames. 
Although the motion in our videos is relatively small, capturing informative content under degraded conditions remains important; therefore, the BRIAR project collects data in the form of video frames.
This, however, introduces significant temporal redundancy between frames. To address this, we adopt a non-learning-based temporal frame selection strategy that efficiently samples a subset of frames while preserving as much of the original motion information as possible.
Figure~\ref{fig:frame_sample} illustrates our sampling strategy designed to reduce computational cost. Specifically, we divide the 30 input frames into four segments. Within each segment, we randomly select two consecutive frames to capture high-frequency jitter caused by turbulence. The resulting 8 frames, aggregated across the four segments, effectively represent the global temporal structure of the video.

\vspace{1mm}
\noindent\textbf{Spatial Region Selection via Learned Attention:}
While temporal selection operates across frames, bottom-up mechanisms also guide focus within a single frame. Here, deep neural networks, particularly those employing self-attention mechanisms such as Transformers~\cite{dosovitskiy2020image}, dynamically compute the relevance of different spatial regions. Each patch in the image contributes to a weighted aggregation of contextual information, enabling the model to emphasize regions with higher relevance based on input features.

\begin{figure}
    \centering
    % \vspace{-5mm}
    \includegraphics[width=.8\linewidth]{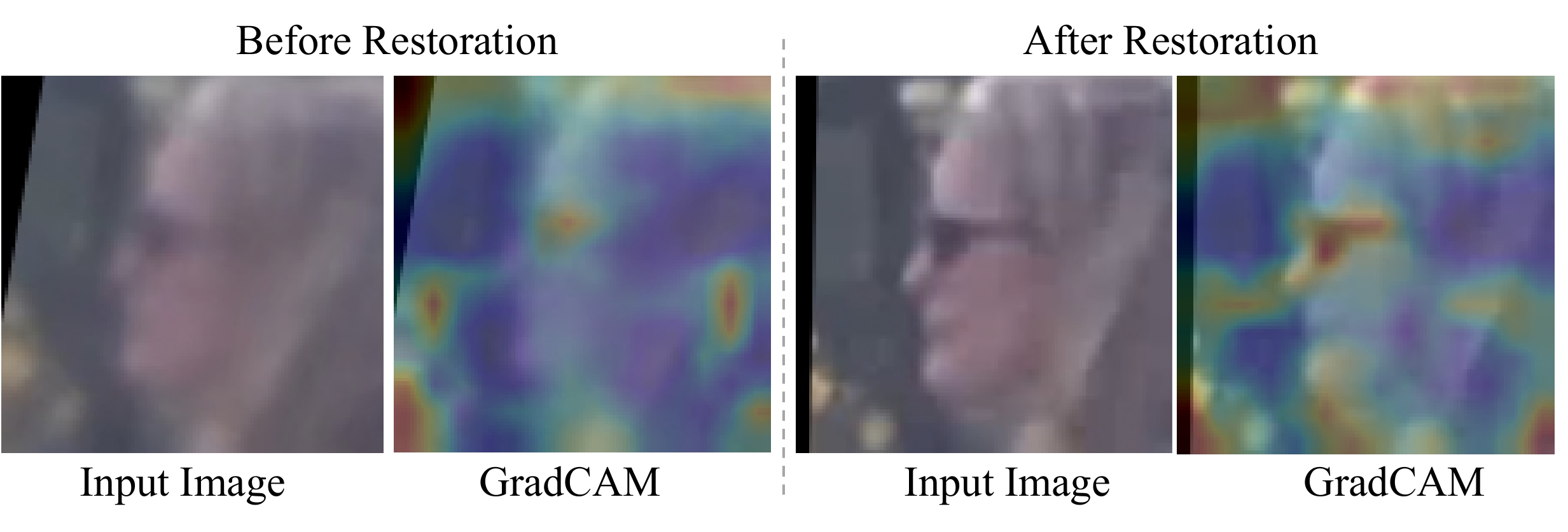}
    % \vspace{-5mm}
    \caption{\revise{Side-by-side comparisons of input images and Grad-CAM visualizations before and after image restoration. With restoration, the input images (left) exhibit noticeable improvements in clarity and detail post-restoration, with sharper features and reduced blur. The Grad-CAM outputs (right) highlight areas of importance for the model's prediction, revealing that the restoration enhances the attention to key regions of interest, particularly around facial features. The alignment of two images is different as images are restored before alignment. \textit{Images shown
with subject permission for publication.}}}
    % \vspace{-3mm}
    \label{fig:spatial_attention}
\end{figure}

This concept of learned spatial attention can also be observed through interpretability techniques like Grad-CAM~\cite{selvaraju2017grad}. As shown in Figure~\ref{fig:spatial_attention}, the Grad-CAM visualizations of the recognition model reveal how restoration improves focus on key features such as facial details. The enhanced clarity and sharpness post-restoration lead the model to allocate greater attention to discriminative regions, improving recognition performance. Unlike temporal fusion mechanisms used for aligning multiple frames, this spatial attention operates within a single frame, highlighting the internal prioritization of visual information by the model.

% Similar spatial attention mechanisms are commonly used within atmospheric restoration models. The temporal fusion often relies on the ``lucky effect'' property of atmospheric turbulence \cite{fried1978probability}. This property refers to a phenomenon in which, during short exposure imaging through atmospheric turbulence, a small subset of frames or image regions are much less distorted than others \cite{french1999catastrophic, aubailly2009automated}. Therefore, the goal of temporal aggregation is to identify and fuse the randomly emerging sharp regions using various spatially varying temporal attention techniques, from the early hand-crafted descriptors or wavelet-based weighting, known as lucky fusion \cite{lau2019restoration, zhu2012removing, anantrasirichai2013atmospheric}, to recent learning based attention \cite{zhang_2024_TMT}. Our restoration model \cite{zhang2024spatio} proposes the temporal fusion. This component is analogous to the classical lucky fusion step, where we implicitly fuse the ``lucky patches'' within the latent feature space, and utilize the recurrent network to integrate the features from previous frames to the current frames.

\subsubsection{Top-down Attention: Learning to Focus from Recognition} 
Even after receiving the selected frames and regions, instance-level selection is still required, as \textbf{not all video clips benefit equally from restoration}.
Empirically, it has been observed that applying restoration to videos that already possess high quality may yield marginal or no improvement for downstream tasks, and can sometimes even degrade performance for very low-quality inputs as shown in Degraded vs. Restore All of Table~\ref{tab: vidcls}.
This may be due to \textit{artifacts introduced during restoration that are inconsistent with recognition-relevant features} \cite{zhang_2024_TMT}.
This motivates a selective approach from top-down attention. For instance, one strategy is to employ a data-driven classifier to predict whether restoration is likely to be beneficial for a given input video, based on its estimated quality and the predicted impact on recognition performance (e.g., rank change). 

\begin{figure}[t]
\centering
% \vspace{-5mm}
\includegraphics[width=1.\linewidth]{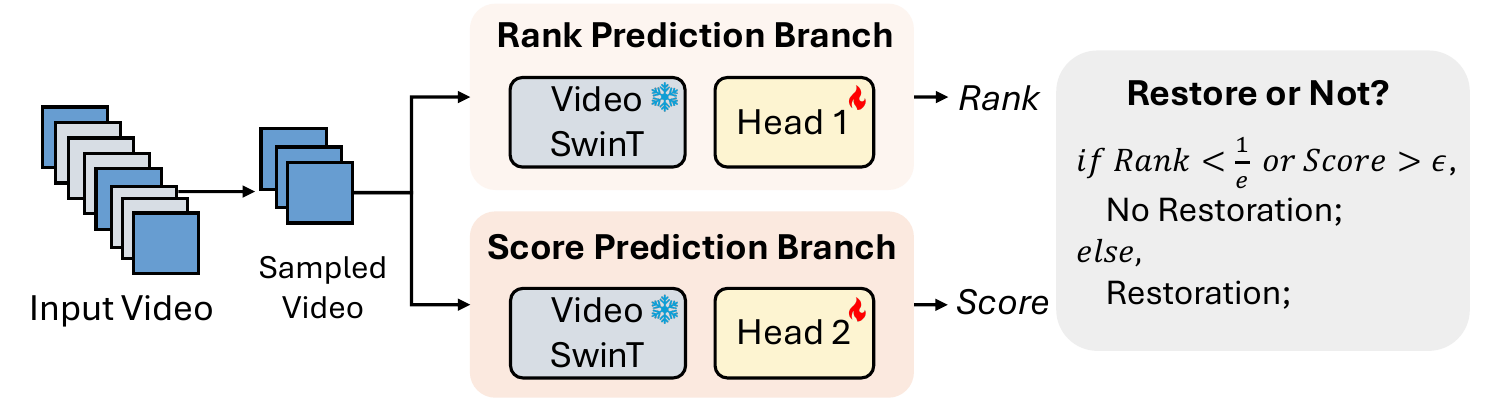}
% \vspace{-3mm}
\caption{\textbf{Video Restoration Classifier (VIDCLS).} The input video is sampled and processed by two branches: Rank Prediction and Score Prediction, both using a pre-trained Video Swin Transformer (SwinT) as the backbone. The decision to restore is based on the predicted rank and score — restoration is skipped if the rank is low or the score is high; otherwise, restoration is applied.}
\label{fig: vidcls_framework}
\end{figure}

\begin{figure}[t]
\centering
% \vspace{-2mm}
\includegraphics[width=0.7\linewidth]{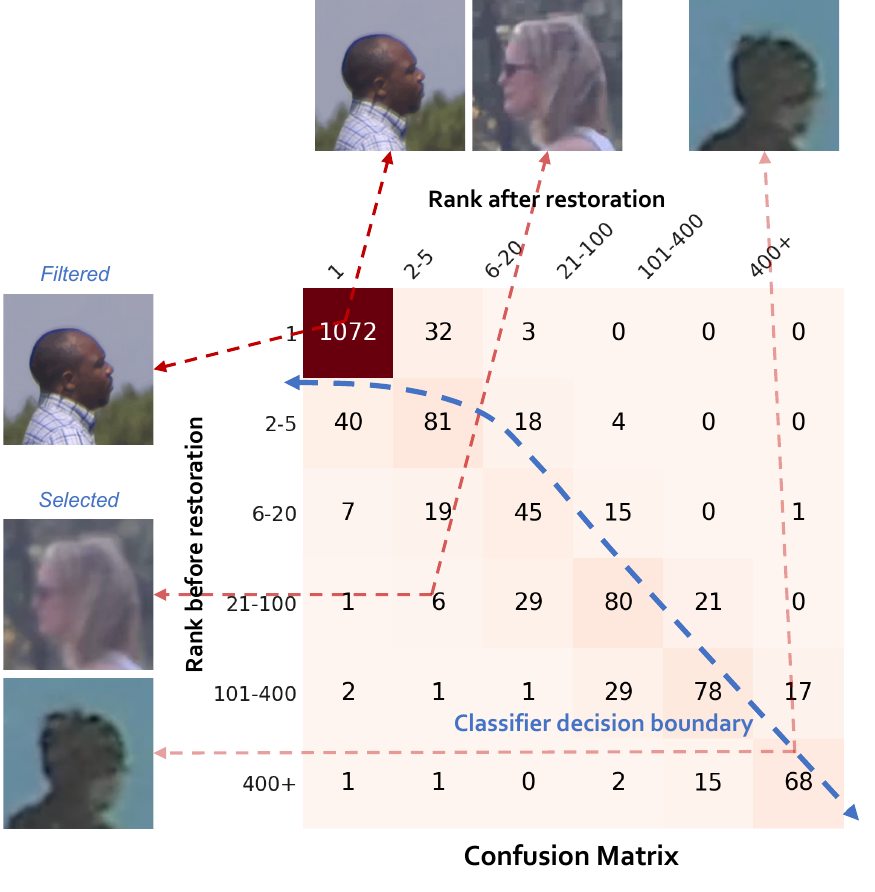}
% \vspace{-3mm}
\caption{The \textbf{confusion matrix and examples} used to train the video restoration classifier. For most videos that already have high quality (Rank 1) before restoration, their rank remains 1 after restoration, indicating that restoration is unnecessary. \textit{Images shown
with subject permission for publication.}}
\label{fig: confusion}
\end{figure}

\noindent\textbf{Recognition-Guided VIDCLS.}
Figure~\ref{fig: vidcls_framework} illustrates the overall architecture of the video restoration classifier (VIDCLS), which consists of two parallel branches: a rank prediction branch and a score prediction branch. The rank prediction branch estimates whether the recognition rank of a video will degrade after restoration, while the score prediction branch assesses whether the input video’s quality is already sufficiently high. Figure~\ref{fig: confusion} presents the confusion matrix and representative examples used to train VIDCLS. The matrix summarizes the rank distributions from our internal BRIAR test set (1,689 videos) before and after restoration, along with sample frames. Notably, approximately 63.5\% of input videos were already rank 1 before restoration and remained so afterward, indicating that restoration is unnecessary for these cases. We also overlay the classifier’s ideal decision boundary in the figure: to the right of this boundary, over 70\% of the videos do not require restoration, which in turn reduces the overall inference latency of the system pipeline and improves the recognition performance.

\begin{table}[t]
    \centering
    \caption{Comparison Before and After Using Video Restoration Classifier, tested on Test Set EVP 4.2.0 Reduced. VIDCLS refers to the proposed Video Restoration Classifier.}
    \renewcommand{\arraystretch}{1.0}
     \setlength{\tabcolsep}{4pt}
    \resizebox{0.9\textwidth}{!}{
    \begin{tabular}{l |c |c |c }
        \toprule
        \textbf{Metric} & \textbf{Degraded} 
 & \textbf{Restore All}& \textbf{VIDCLS} \\
        \midrule
        \multirow{1}{*}{1:1 TAR@0.1\%FAR$\uparrow$} 
        & 62.6\%  & \textbf{63.5\%} & 63.3\% \\
        % \midrule
        \multirow{1}{*}{1:N Rank Top 20$\uparrow$} 
         & \textbf{87.2\%}  & 86.5\% & 86.9\% \\
        % \midrule
        \multirow{1}{*}{1:N Search FNIR@1\%FPIR$\downarrow$} 
         & 51.2\%  & 51.6\% &\textbf{49.9\%} \\
         \midrule
         $\Delta$ vs. Degraded & -- & $-$0.07\% & $+$0.57\%\\
         % \midrule
         % Inference Time & \\
        \bottomrule
    \end{tabular}}
    \label{tab: vidcls}
\end{table}

\vspace{1mm}
\noindent\textbf{Recognition-Relevant Rank Label.}
During training, the video Swin Transformer remains frozen, and only the two heads are trained separately. The two branches use different training strategies. 
For the video quality prediction branch, we directly use the similarity score as a proxy for video quality. During inference, after processing the video through the two branches, if the predicted video quality exceeds a certain threshold or the predicted rank is expected to decrease, restoration will not be applied to that video.
For the rank change prediction branch, we considered the initial rank \(\text{rank}_{\text{raw}}\) of the video when assigning its label. The final rank label is defined as follows,
\begin{equation}
\text{label}_{\text{rank}} = \exp\left(\frac{\text{rank}_{\text{raw}} - \text{rank}_{\text{post}}}{\text{rank}_{\text{raw}}} - 1\right)
\end{equation}
 where \(\text{rank}_{\text{post}}\) represents the rank after applying restoration. This formula constrains each video's label to the range [0,1], with larger values indicating a more significant rank improvement after restoration. A value of \(1/e\) indicates that the rank remains unchanged. 

\vspace{1mm}
\noindent\textbf{Re-Weighted to Address Label Imbalance.}
When training the classifier, we found that due to the extremely imbalanced data labels (most videos have a rank of 1, which does not change), the network struggles to optimize the parameters. Therefore, we automatically adjust the loss function's weight based on the rank label of the input video. The weight adjustment function is given by:
\begin{align}
\nonumber f(m) = & -1.25 + \mathcal{N}\left(m, -0.25, 0.2^2\right) \\
& + \mathcal{N}\left(m, -\frac{1}{e}, 0.18^2\right) + \mathcal{N}\left(m, 1 - \frac{1}{e}, 0.3^2\right)
\end{align}
where $m = \text{label}_{\text{rank}} - \frac{1}{e}$. By applying this weight, we force the network to focus more on videos with more significant rank changes during training.

As demonstrated in Table \ref{tab: vidcls}, using such a VIDCLS to selectively skip the restoration process for certain videos led to an overall improvement in face recognition accuracy compared to applying no restoration, while also significantly reducing computational overhead by avoiding unnecessary processing. This video-level filtering acts as a form of top-down attention, deciding whether to allocate processing resources based on the top-recognition task.

% (visualize attention basically, would be better if we could find examples to show “over-restoring” clean frames will damage recog)

% Our observation is restoration does \emph{not} always improve the recognition, even if the image quality is significantly improved. \fref{fig:example}(a) shows an example of pairing a restoration module GRTM  \cite{liu2024farsight} with a recognition module AdaFace \cite{kim2022adaface}. The two modules are individually trained where the restoration aims to improve the image quality, and the recognition model aims to recognize the face. Both models used synthesized turbulence data as the training set. There are two noticeable observations: Image quality improves, but similarity score drops. So better restoration does not translate to better recognition.

% Therefore, inspired by the human vision system, it is vitally important to ensure that restoration efforts are directly aligned with the top recognition objectives, following a top-down attention mechanism. To fulfill this goal, we adopt a co-training pipeline for restoration and recognition (Figure \ref{fig:co-training}), which is illustrated in detail in Section~\ref{sec:computer_vision_angel}.

\subsection{Selective Neuroplasticity}
% Building on the concept of \textbf{neuroplasticity}, the entire pipeline can be adaptively fine-tuned by selectively updating specific components to address degradation-specific challenges. This targeted adaptation enables the model to reconfigure its internal representations in response to different types of input corruption. To systematically explore this capability, we conducted extensive experiments to assess the impact of fine-tuning different stages of the restoration-recognition pipeline — namely, the early, middle, and late layers. The results provide insights into which parts of the network are most sensitive or critical for maintaining performance under various degradation conditions.

% \begin{itemize}
%     \item \textbf{Fine-Tuning Early Layers.} Early layers of the restoration module focus on low-level feature extraction, making them pivotal for handling degradations such as noise and blur. For instance, early layer adjustments can better preserve high-frequency details.
%     \item \textbf{Fine-Tuning Middle Layers.} Middle layers play a critical role in interacting with degradation-specific priors, such as those informed by physics-based models. These layers were most effective in handling complex atmospheric distortions.
%     \item \textbf{Fine-Tuning Late Layers.} Late layers are responsible for high-level semantic feature alignment, particularly identity-preserving features in biometrics. These layers were found to be crucial for restoring identity-related details in face recognition.
% \end{itemize}

Building on the concept of \textbf{neuroplasticity}, the entire pipeline can be adaptively fine-tuned by selectively updating specific components to address degradation-specific challenges more effectively. This targeted adaptation mechanism empowers the model to dynamically reconfigure its internal representations in response to diverse types of input corruption, thereby enhancing its robustness. 
\revise{To further investigate the adaptive capacity of the pipeline, we conducted a comparative study by selectively fine-tuning specific components. This analysis is motivated by the well-established understanding that deep neural networks exhibit a clear functional hierarchy \cite{zeiler2014visualizing,yosinski2014transferable}: early layers typically capture general low-level textures, while late layers encode task-specific semantic abstractions.}

\begin{itemize}
    \item \textbf{Fine-Tuning Early Layers.} Early layers of the restoration module focus on low-level feature extraction, making them pivotal for handling degradations such as noise and blur. For instance, early layer adjustments can better preserve high-frequency details.
    \item \textbf{Fine-Tuning Middle Layers.} Middle layers play a critical role in interacting with degradation-specific priors, such as those informed by physics-based models. These layers were most effective in handling complex atmospheric distortions.
    \item \textbf{Fine-Tuning Late Layers.} Late layers are responsible for high-level semantic feature alignment, particularly identity-preserving features in biometrics. These layers were found to be crucial for restoring identity-related details in face recognition.
\end{itemize}

\begin{table}[h]
    \centering
        \caption{Face recognition metrics when fine-tuning different components of the restoration network with co-training on a BRS validation set.}
        \renewcommand{\arraystretch}{1.1}
    \begin{tabular}{l| c}
    \toprule
        \textbf{Trainable Layers} &  \textbf{Validation Accuracy} \\
        \midrule
         Early Layers & 73.22\%\\
         Middle Layers & 73.38\%\\
         Late Layers & 73.54\%\\
         \midrule
         All Layers & \textbf{73.62\%} \\
         \bottomrule
    \end{tabular}
% \vspace{-2mm}
    \label{tab:different_layer}
\end{table}

\revise{As summarized in Table~\ref{tab:different_layer}, the results indicate that fine-tuning all layers simultaneously yields the optimal performance, as it allows for a holistic adaptation to the degradation. However, when examining partial fine-tuning configurations, we observe that updating the late layers leads to a relatively higher validation accuracy (73.54\%) compared to the early (73.22\%) or middle stages (73.38\%). This trend aligns with the findings in \cite{zhang2018unreasonable}, suggesting that for downstream recognition tasks, maintaining high-level semantic consistency is often more critical than recovering low-level pixel-wise fidelity. Our results confirm that while all levels contribute to the final restoration quality, the late-stage alignment serves as a more efficient bottleneck for bridging the gap between restoration and recognition.}

% As shown in Table~\ref{tab:different_layer}, tuning the late layers yields the best performance, comparable to tuning all layers. This suggests that semantic information serves as a crucial bridge between the restoration and recognition modules.
% This indicates that alignment between the two components is most effectively achieved at higher semantic levels, where abstract and task-relevant features are captured, rather than at lower levels focused on raw signal fidelity. Nevertheless, the shallow and middle layers also contribute to performance, indicating that multi-level adaptation can still provide benefits. This design enables more effective task-aware restoration by guiding optimization with recognition objectives while preserving the core image reconstruction process.

% Consequently, it becomes beneficial to partially preserve the early and middle components of the restoration module, allowing the network to retain its general low-level enhancement capabilities while concentrating adaptation efforts on the high-level semantic representations. 

\begin{tcolorbox}[before skip=2mm, after skip=0.0cm, boxsep=0.0cm, middle=0.0cm, top=0.1cm, bottom=0.1cm]
    \textit{\textbf{Takeaways:}\\
% \begin{itemize}[leftmargin=*]
    \textit{\textbf{(A6)}} \textbf{“Less is more”}: Sampling frames, cropping focused regions, skipping videos not only improves efficiency and also enhance recognition accuracy.\\
    \textit{\textbf{(A7)}} \textbf{Tuning late layers wins over tuning others}:
Semantic information serves as the bridge between recognition and restoration.
    % Partial preservation of restoration is essential, with most adaptation occurring at ``high semantic level'' rather than ``low signal level''.
% \end{itemize}
}
\end{tcolorbox}

\section{Evaluation Protocol}\label{sec:evaluation}

\noindent\textbf{Real World Degradation Training Dataset.}
We utilize the BRIAR dataset~\cite{briar} from the IARPA BRIAR project to study the effect of restoration on recognition in the presence of real world image degradation. BRIAR dataset includes both government-collected datasets and datasets contributed by program participants: Accenture, KITWARE, MSU, USC, and STR. A summary of the training dataset statistics is provided in Table~\ref{tab:briardata}. 
% Example images are shown in Fig~\ref{fig: Example Rank}. 
Face recognition model is initially pre-trained on WebFace12M dataset~\cite{zhu2021webface260m} and subsequently fine-tuned on BRIAR dataset. All subjects in this paper provided consent for their images to be published.

\vspace{1mm}
\noindent\textbf{Face Recognition Testing Dataset.}
The testing dataset is collected as part of the BRIAR program and comprises 424 probe and gallery subjects, along with 675 additional distractor identities. In total, the dataset contains 362,210 video frames for the probes and 8,214,485 frames for the gallery set. We follow the \textit{Face Included Treated} protocol, which focuses on visible faces captured under challenging conditions such as long-range distances, elevated viewing angles, and unmanned aerial vehicle (UAV) platforms.

\vspace{1mm}
\noindent\textbf{Face Recognition Evaluation Metrics.}
To assess recognition performance, we utilize the BRIAR Program Target Metrics~\cite{briar,liu2024farsight}, which provide a comprehensive evaluation of biometric recognition. Our evaluation framework focuses on three critical metrics: 1:1 verification, closed-set identification, and open-set identification, each of which captures different aspects of biometric recognition performance.
1:1 verification measures the ability to correctly verify an individual's identity by comparing a probe image to a reference template, using the True Acceptance Rate (TAR) at a 0.01\% False Acceptance Rate (FAR). Closed-set identification, on the other hand, assesses how accurately the system can recognize an individual from a known database by measuring Rank-20 accuracy. Open-set identification evaluates the system’s ability to distinguish whether a probe belongs to an enrolled subject or a distractor by calculating the False Non-Identification Rate (FNIR) at a 1\% False Positive Identification Rate (FPIR). 

\begin{table}[t]
\centering
\small
\caption{Training Dataset Statistics from BRIAR dataset.}
\renewcommand{\arraystretch}{1.0}
% \resizebox{0.4\textwidth}{!}{
\begin{tabular}{@{}lcc@{}}
\toprule
\textbf{Dataset}      & \textbf{\# Subjects} & \textbf{\# Media (videos/images)} \\ \midrule 
BRIAR-BRC    & 995        & 162,949                  \\
MSU-BRC      & 452        & 5,599/17,593             \\
Accenture-BRC & 512       & 21,948/21,204            \\
Kitware-BRC  & 509        & 568,263                  \\
USC-BRC      & 290        & 26,222                   \\
STR-BRC      & 436        & 8,394/25,135             \\ 
\bottomrule
\end{tabular}
\label{tab:briardata}
\end{table}

\vspace{1mm}
\noindent\textbf{Visual Quality Evaluation Metrics.}
In addition to recognition-based evaluation, we assess image quality in selected experiments to analyze the gap between visual and recognition metrics. Since PSNR and SSIM require high-quality references, we create an evaluation set by manually selecting 1,782 clean face images (180 identities) from the BRIAR gallery and simulating atmospheric turbulence to generate degraded-clean pairs. Results are shown in Figure~\ref{fig:different_lambda} and Table~\ref{tab:different_weight}.

\section{Conclusion and Future Directions}\label{sec:conclusion}

In this work, we proposed an interdisciplinary framework that integrates neuroscience-inspired mechanisms, physics-driven signal processing, and computer vision techniques to enhance machine recognition in degraded conditions. By co-optimizing restoration and recognition tasks, we demonstrated that restoration pipelines can be designed not just for perceptual quality but to actively improve downstream recognition performance. Our feedback-driven loops, inspired by neuroplasticity, dynamically adapt restoration parameters based on recognition outputs, enabling robust and adaptable recognition across diverse scenarios.
Through extensive experimentation on challenging benchmarks BRIAR, we validated the effectiveness of our approach. Key findings include:
\begin{itemize}
\item \textbf{Physics-Driven Integration}: Embedding physics-based priors, such as turbulence simulations, into restoration models enhances generalization to real-world conditions.
    \item \textbf{Task-Specific Restoration}: Aligning restoration with recognition objectives significantly improves biometric and multimodal recognition accuracy, particularly in turbulence and noise-degraded settings.
    \item \textbf{Neuroplasticity-Inspired Selection}: Integrating attention mechanisms across frames, spatial regions, and entire video sequences enhances both computational efficiency and recognition performance.
    % \item \textbf{Neuroplasticity-Inspired Selection}: Targeted adjustments to specific parts of the pipeline enable precise handling of complex degradations, demonstrating the value of adaptive and feedback-driven strategies.
\end{itemize}
Despite these advances, challenges remain. The computational demands of integrating multiple methodologies, the need for better task-specific evaluation protocols, and the limitations of synthetic training data highlight areas for further innovation.

\vspace{1mm}
\noindent\textbf{Future Directions.}
Looking ahead, expanding the scope of neuroscience-inspired mechanisms offers an exciting avenue for advancing restoration-recognition synergy. While this work explored hierarchical attention and feedback-driven loops inspired by predictive coding, future frameworks could delve deeper into biological principles such as cortical plasticity, where dynamic adaptation not only adjusts parameters but also alters the architecture based on the nature of degradations. This adaptability could be further enhanced by leveraging multisensory integration, combining visual cues with auxiliary modalities like audio or tactile inputs to reinforce recognition under extreme visual impairments. These advancements promise to deepen the connection between biological intelligence and artificial systems, enabling recognition models to operate robustly in unpredictable conditions. 

Additionally, improved modeling of complex degradations, particularly those combining multiple environmental factors, remains a critical challenge. Incorporating real-world variability, such as dynamic lighting, turbulence, and motion, into physics-based priors and synthetic training data could significantly enhance generalization. Extending these efforts to unseen domains, including medical imaging and environmental monitoring, presents an opportunity to broaden the applicability of restoration-recognition frameworks.

Efficiency remains another key limitation as interdisciplinary integration introduces computational overhead. Addressing this gap requires optimization techniques that balance computational complexity with performance, such as hardware-aware models designed for edge devices or hybrid neuro-symbolic systems that leverage symbolic reasoning to reduce resource demands. Finally, ethical considerations and fairness must guide the development of restoration-recognition systems, especially in biometric applications. Restoration pipelines could inadvertently introduce biases or amplify disparities across demographic groups. Ensuring fairness-aware algorithms that maintain equitable performance across all populations is essential for ethical deployment in real-world applications. Together, these directions highlight the challenges and opportunities in building robust, adaptable, and equitable recognition systems that meet the demands of diverse and complex scenarios.

% \clearpage

\vspace{1mm}
\noindent\textbf{Acknowledgments.} This research is based upon work
supported in part by the Office of the Director of National Intelligence (ODNI), Intelligence Advanced Research
Projects Activity (IARPA), via 2022-21102100004. The views
and conclusions contained herein are those of the authors
and should not be interpreted as necessarily representing
the official policies, either expressed or implied, of ODNI,
IARPA, or the U.S. Government. The U.S. Government is
authorized to reproduce and distribute reprints for governmental purposes notwithstanding any copyright annotation
therein.

%BACKMATTER SEE DOCUMENTATION
\printbibliography

\end{document}